\documentclass{ieeeaccess}

\usepackage{booktabs}       
\usepackage{nicefrac}       
\usepackage{graphicx}       
\usepackage{graphicx}%
\usepackage{multirow}%
\usepackage{amsmath,amssymb,amsfonts}%
\usepackage{amsthm}%
\usepackage{mathrsfs}%
\usepackage{xcolor} 
\usepackage{soul}
\usepackage{textcomp}%
\usepackage{manyfoot}%
\usepackage{algorithm}%
\usepackage{algorithmicx}%
\usepackage{algpseudocode}%
\usepackage{listings}%
\usepackage{array}
\usepackage{comment}
\usepackage{acronym}
\usepackage{xspace}
\usepackage[mathscr]{eucal}
\usepackage[export]{adjustbox}
\DeclareMathOperator*{\argmax}{arg\,max}
\usepackage{subfig}
\usepackage{longtable}

\newcommand*{\eg}		{e.g.,\ }
\newcommand*{\ie}		{i.e.,\ }

\acrodef{ML} 		[\textsc{ML\xspace}]				{Machine Learning}
\acrodef{DL} 		[\textsc{DL\xspace}]				{Deep Learning}
\acrodef{DNN} 		[\textsc{DNN\xspace}]				{Deep Neural Network}
\acrodef{CNNs} 		[\textsc{CNNs\xspace}]				{Convolutional Neural Networks}
\acrodef{CNN} 		[\textsc{CNN\xspace}]				{Convolutional Neural Network}
\acrodef{MTL} 		[\textsc{MTL\xspace}]				{Multi-Task Learning}
\acrodef{RL} 		[\textsc{RL\xspace}]				{Reinforcement Learning}
\acrodef{LL} 		[\textsc{LL\xspace}]				{Lifelong Learning}
\acrodef{NLP} 		[\textsc{NLP\xspace}]				{Natural Language Processing}
\acrodef{LSTM} 		[\textsc{LSTM\xspace}]				{Long Short-Term Memory}
\acrodef{MAML} 		[\textsc{MAML\xspace}]				{Model Agnostic meta-learning}
\acrodef{SMR} 		[\textsc{SMR\xspace}]				{Structured Multi-output Regression}
\acrodef{NER} 		[\textsc{NER\xspace}]				{Named Entity Recognition}
\acrodef{EMR} 		[\textsc{EMR\xspace}]				{Electronic Medical Records}
\acrodef{POS} 		[\textsc{POS\xspace}]				{Parts-of-Speech}
\acrodef{MMMTL} 	[\textsc{$\mathrm{M^3TL}$\xspace}]	{Multi-modal multi-task meta transfer learning}
\acrodef{ADME}  	[\textsc{ADME\xspace}]	            {Absorption, Distribution, Metabolism, and Excretion}
\acrodef{GAN}   	[\textsc{GANs\xspace}]	               {Generative Adversarial Nets}
\acrodef{VAE}   	[\textsc{VAEs\xspace}]	               {Variational Auto-Encoders}

\usepackage{hyperref}
\usepackage{url}            

\usepackage{bm}
\makeatletter
\AtBeginDocument{\DeclareMathVersion{bold}
\SetSymbolFont{operators}{bold}{T1}{times}{b}{n}
\SetSymbolFont{NewLetters}{bold}{T1}{times}{b}{it}
\SetMathAlphabet{\mathrm}{bold}{T1}{times}{b}{n}
\SetMathAlphabet{\mathit}{bold}{T1}{times}{b}{it}
\SetMathAlphabet{\mathbf}{bold}{T1}{times}{b}{n}
\SetMathAlphabet{\mathtt}{bold}{OT1}{pcr}{b}{n}
\SetSymbolFont{symbols}{bold}{OMS}{cmsy}{b}{n}
\renewcommand\boldmath{\@nomath\boldmath\mathversion{bold}}}
\makeatother

\def\BibTeX{{\rm B\kern-.05em{\sc i\kern-.025em b}\kern-.08em
    T\kern-.1667em\lower.7ex\hbox{E}\kern-.125emX}}

\begin{document}
\history{Received 12 August 2024, accepted 7 October 2024, Date of publication xxxx 00, 0000, date of current version  14 October 2024.}
\doi{10.1109/ACCESS.2024.3478805}

\title{Sharing to learn and learning to share;
Fitting together Meta, Multi-Task, and Transfer Learning: A meta review}
\author{Richa Upadhyay\authorrefmark{1}, 
Ronald Phlypo\authorrefmark{2}, Rajkumar Saini\authorrefmark{1}, and Marcus Liwicki\authorrefmark{1}
\IEEEmembership{Senior Member, IEEE}}

\address[1]{Lule\aa ~ University of Technology, Lule\aa, Sweden (e-mail: firstname.lastname@ltu.se)}
\address[2]{GIPSA Lab, Grenoble INP, Grenoble,
France (e-mail: ronald.phlypo@grenoble-inp.fr)}

\markboth
{Upadhyay \headeretal: Sharing to learn and learning to share}
{Upadhyay \headeretal: Sharing to learn and learning to share}

\corresp{Corresponding author: Richa Upadhyay (e-mail: richa.upadhyay@ltu.se).}

\begin{abstract}
Integrating knowledge across different domains is an essential feature of human learning. 
Learning paradigms such as transfer learning, meta-learning, and multi-task learning reflect the human learning process by exploiting the prior knowledge for new tasks, encouraging faster learning and good generalization for new tasks.
This article gives a detailed view of these learning paradigms and their comparative analysis.
The weakness of one learning algorithm turns out to be a strength of another, and thus, merging them is a prevalent trait in the literature.
Numerous research papers focus on each of these learning paradigms separately and provide a comprehensive overview of them.
However, this article reviews research studies that combine (two of) these learning algorithms.
This survey describes how these techniques are combined to solve problems in many different fields of research, including computer vision, natural language processing, hyper-spectral imaging, and many more, in a supervised setting only.
Based on the knowledge accumulated from the literature, we hypothesize a generic task-agnostic and model-agnostic learning network – an ensemble of meta-learning, transfer learning, and multi-task learning, termed Multi-modal Multi-task Meta Transfer Learning.
We also present some open research questions, limitations, and future research directions for this proposed network.
The aim of this article is to spark interest among scholars in effectively merging existing learning algorithms with the intention of advancing research in this field.
Instead of presenting experimental results, we invite readers to explore and contemplate techniques for merging algorithms while navigating through their limitations.
\end{abstract}


\titlepgskip=-21pt

\maketitle

\section{Introduction} \label{Introduction}
\ac{ML} continuously draws inspiration from human cognition and decision-making to develop more human-like, neurally-weighted algorithms  \cite{fong2017using}. 
Traditional \ac{ML} algorithms follow a single-task learning approach wherein they are trained to solve only one task at a time.
If there is a need to accomplish another task, the network must be re-trained on a new dataset. 
single-task learning is also often referred to as \textit{isolated learning}  \cite{LL_isolated}.
It does not utilize and preserve any prior information from the previous learning for a future task.
However, humans do not learn anything from scratch, and they can rapidly learn new concepts due to their inherent potential to share the acquired knowledge across tasks seamlessly. 
The more related the tasks, the easier it is to utilize the knowledge.
For example, the knowledge acquired while riding a bike or a motorbike is useful when learning to drive a car. 
The transfer of prior knowledge enables humans to learn quickly and accurately in a few instances because humans are biased that similar tasks have similar solutions (to some extent). 
Therefore, they acquire the concept by focusing on learning the differences between the features of the tasks.
Some of the learning techniques in \ac{ML} or \ac{DL}, such as transfer learning  \cite{GoodBengCour16, Pan10asurvey}, meta-learning  \cite{Thrun98,baxter1998theoretical, vettoruzzo2024advances}, \ac{MTL}  \cite{MTL, Crawshaw2020, vandenhende2021multi, yu2024unleashing}, and lifelong learning  \cite{LL_book,soltoggio2024collective}, are inspired by such human capability, where the aim is to transfer the learning of one task to another.
These algorithms are often referred to as knowledge-sharing algorithms.

\subsection{Knowledge transfer in machine learning algorithms}
Learning one task at a time is a generic approach in the field of \ac{ML}.
Big problems are disassociated into smaller independent sub-tasks that are learned distinctly, and combined results are presented.
The concept of \ac{MTL} is introduced by  \cite{MTL}, who proposed that all tasks should be trained simultaneously to achieve better performance. 
The underlying concept is that if all the sub-tasks share their learning, they may find it easier to learn rather than in isolation.  
It is inspired by the concept of distributed representations in the human brain, in which a task is divided into smaller sub-tasks, and different neural networks represent different sub-tasks in different regions of the brain  \cite{yang2019study}. 
Still, these networks share common features and representations.

Similarly, learning a complex task such as riding a bike relies on motor skills that humans develop when they learn to walk as toddlers.
The task is, therefore, learned by integrating a considerable amount of prior knowledge across tasks.
Additionally, while learning generalized concepts across many tasks, humans evolve the ability to learn quickly and in fewer instances. 
These concepts are the bedrock for \textit{transfer learning}  \cite{GoodBengCour16,Pan10asurvey}, \textit{lifelong learning}  \cite{LL_book,soltoggio2024collective}, and \textit{meta-learning}  \cite{Thrun98,baxter1998theoretical, vettoruzzo2024advances}.
The approach of `how' prior knowledge is introduced while training makes these learning algorithms mutually distinct.

\subsection{Types of knowledge transfer} \label{negative_transfer}

Information sharing between different tasks can result in two types of knowledge transfers: positive and negative  \cite{Crawshaw2020, zhang2021survey}. 
Positive transfer occurs when the information shared between the tasks aids in improving the performance of the tasks.
In contrast, the negative transfer occurs when the performance of the tasks suffers due to information flow within the tasks.
Negative transfer is also known as destructive interference; it occurs because possibly even related tasks may have contradictory requirements.
When these are in a knowledge-sharing setting, improving the performance of one task may hinder the performance of another. 
Negative transfer is a significant issue in multiple-task learning concepts such as transfer learning, \ac{MTL}, etc.
To support positive transfer, it is essential for an algorithm to fulfill two complementary objectives: retain task-specific knowledge and better generalization across tasks.
Accomplishing such goals is very complicated; consequently, this is an active area of research.

\subsection{What is a Task?} \label{sec:task}
Before discussing the knowledge-sharing learning paradigms, it is essential to define what is a `task.'
While discussing learning paradigms in this article, it should be noted that the task does not address the learning process.
In fact, learning assists in acquiring the knowledge necessary to perform a task.
A task can formally be defined as \textit{work done to achieve an objective}.
Regression, segmentation, machine translation, classification, dimension reduction, and anomaly detection are the most typical tasks performed by various \ac{DL} algorithms.
In the broader concept of \ac{ML}, a 'task' can typically encompass the entire process from data acquisition through learning to decision making; we make a clear distinction between tasks and the learning process itself.
In the context of this paper, a task specifically refers to the goal (e.g., classification or segmentation) that the learning methods aim to achieve. 
We believe this differentiation helps to better separate the tasks from the methods (or learning paradigms) employed to accomplish them.

Two tasks are considered homogeneous if they share a similar but non-identical objective and heterogeneous if they share a different objective.
For example, an image-level classification task and a semantic segmentation task, \ie pixel-level classification, are considered heterogeneous (or dis-similar or different) tasks.
In contrast, two image classification tasks are considered homogeneous (or similar).
The formal definition of a task is further explained in detail in Sec. \ref{supervised_learning}.

Note that the terms task and domain are considered different in this article.
Researchers interchangeably use these terms when there is only one task over each domain.
However, in this article, domain refers to the data distribution from which the training or test data are sampled, the exact definition in Sec. \ref{supervised_learning}.
In general, the tasks in supervised learning can be grouped according to the labels or ground truth. \eg if the labels are categorical, the task is classification; if they are continuous, the task is regression; and if the labels are categorical at the pixel level, the task is segmentation.
While the domain is connected to the input feature space, \eg images of handwritten numbers (such as MNIST,  \cite{deng2012mnist}) come from a different domain than images of printed house numbers (such as SVHN,  \cite{netzer2011reading}).


\subsection{Scope of this work}
The emphasis of this article is on algorithms that share information between tasks to improve their learning.
Particularly, three learning paradigms are discussed: multi-task learning, meta-learning, and transfer learning.
Other related learning algorithms, such as \ac{LL}, online learning, and \ac{RL}, are not within the scope of this work because their fundamental objective is slightly different, \ie the progressive learning process. 
Additionally, reinforcement learning and online learning are usually restricted to single-task learning, while the others involve multiple tasks.
\ac{LL} has the key characteristics of consistent knowledge accumulation across several tasks and reusing it while learning a new task, which makes it closely related to meta-learning.
However, the system architecture of \ac{LL}  \cite{LL_book} is fundamentally different from the other learning paradigms: it may require an ensemble of many learning algorithms and various knowledge representation methods. 
In order to evaluate \ac{LL}, many tasks and datasets are required to review the algorithm's performance, while learning the sequence of tasks is of great significance.
Given these differences, \ac{LL} is excluded from this study.

Although this paper provides an in-depth explanation of multitask, meta, and transfer learning, it \textit{does not} discuss the literature devoted exclusively to these algorithms. 
This is due to the fact that the central purpose of this work is to talk about the research articles that focus on combining these learning paradigms.
Numerous scholarly articles provide a comprehensive survey of these distinct algorithms and discuss recent developments in the relevant fields, like  \cite{Crawshaw2020, Ruder2017AnOO, zhang2018overview, vandenhende2021multi, thung2018brief, chen2021multi, 9706456, yu2024unleashing} discuss \ac{MTL},  \cite{meta_survey, peng2020comprehensive, luo2022meta, lee2022meta, vilalta2002perspective, huisman2021survey, vettoruzzo2024advances} gives survey on meta-learning algorithms, and  \cite{9134370,TTL,weiss2016survey,Pan10asurvey, 9336290,9789336} review transfer learning.
The fact that there is no such comprehensive review of works that have combined these learning algorithms is the motivation behind our presentation of a survey of research papers that fall under this category. 

\subsection{Contribution}
The prime focus of this article is learning paradigms that can handle multiple tasks and transfer information between those tasks, particularly \ac{MTL}, meta-learning, and transfer learning. 
Even though they all are knowledge-sharing algorithms and deal with multiple tasks, they differ in why information is shared, what information is shared, and when it is shared.
For example, in \ac{MTL}, the tasks are trained jointly and exchange knowledge as model parameters. 
In contrast, in meta-learning, several tasks are trained sequentially and accumulate meta-knowledge across tasks, which can be the parameters, hyper-parameters, model architecture, and so on; the meta-knowledge is subsequently applied to the new task.
On the other hand, transfer learning aims to use what has already been learned in one setting (or domain) to enhance learning in another situation (or domain).
The main distinction between these algorithms is their primary goal; some focus on improving performance by information sharing, while others aim at faster learning, better generalization, or stable learning.
The following are the significant contributions of this work.
\begin{enumerate}

    \item An outline of the three learning algorithms, \ie transfer learning, \ac{MTL}, and meta-learning, together with a comparative study of these learning paradigms focusing on their strengths and weaknesses. 
    \item A detailed survey of the current research in multi-task meta-learning, meta transfer learning, and multi-task transfer learning.  
    This article highlights instances in the literature where these algorithms are frequently misunderstood as similar or derivatives of one another due to the misuse of terminology.
    In addition, Table \ref{tab:datasets} provides a summary of the tasks and corresponding datasets for all previous studies discussed in this article. 
    \item With the insights from the literature, this article puts together a generic task-agnostic and model-agnostic learning network (a hypothesis), a method for combining these three learning paradigms in a way that allows for the use of all three learning algorithms, combinations of two, or just one of them as needed.
    \item This work discusses open questions related to knowledge sharing in the ensemble of \ac{MTL}, meta-learning, and transfer learning. 
\end{enumerate}

The title of this paper, "Sharing to learn and learning to share," has significance to the learning paradigms explored in the paper, namely \ac{MTL}, meta-learning, and transfer learning. 
\textit{Sharing to learn} is a direct reference to \ac{MTL}, where multiple tasks share representations or parameters to mutually enhance their learning outcomes.
And, \textit{learning to share} refers to the core idea of meta-learning, where the learning algorithm itself learns how to optimize the sharing of knowledge across different tasks.
As for transfer learning, while its main goal is the transfer of knowledge from one domain/task to another, this algorithm indirectly supports the idea of "learning to share" by leveraging pre-learned knowledge to improve performance on new tasks.

Regarding the exclusions of this article, it \textit{does not} delve into any experimental results, either from the surveyed literature or our own research. This omission primarily stems from the incompatibility of comparing the performance across the cited articles, given the disparate datasets and diverse settings employed in each work.
Besides, the central aim of this paper is to inspire the research community to investigate methods for proficiently merging these knowledge-sharing algorithms with a wide-reaching goal of improving generalization, learning using less data, faster learning, and ensuring robust performance. 


\subsection{Structure}
The structure of the article is as follows: Sec.~\ref{learning_paradigms} gives the definitions and fundamentals of the learning paradigms along with examples and a comparative study.
A detailed literature study of the learning algorithms when used together is presented in Sec.~\ref{ensemble}.
Sec.~\ref{discussion} introduces a generic learning network, which is an ensemble of the three learning algorithms, and lists a few open research questions related to it.
Some of the important limitations of the ensemble are also discussed in this section. 
Finally, Sec.~\ref{questions} concludes the article and gives directions for future research. 


\section{Learning paradigms} \label{learning_paradigms}

This section elaborates on three learning paradigms highlighted in this article: transfer learning, \ac{MTL}, and meta-learning. 
In addition to a detailed definition, a comparison of these algorithms is provided below. 
However, this is preceded by a discussion of conventional supervised learning and its limitations, which explain the need for an information-sharing algorithm.

\subsection{Definitions and Notations}

Before getting into the discussions about the learning paradigms, it is important to know some of the notations used in this article. These are:
\begin{itemize}
    \item Variables in calligraphic ($\mathcal{D,X,Y,T}$) represent sample space, particularly -
    \begin{itemize}
        \item $\mathcal{D}$ : stand for domain
        \item $\mathcal{X, Y}$: represent the input feature space and output label space, respectively
        \item $\mathcal{T}$ : is the task space 
    \end{itemize}
    \item Variables in italics ($D,x,y,T$) represent one sample of the space
    \begin{itemize}
        \item $D$ : denotes a dataset
        \item $x,y$: stands for an input and output sample, respectively
        \item $T$ : represents a task
        \item $L$ : is the loss function
    \end{itemize}
    \item $p(A,B)$ - joint probability distribution of A and B
    \item $p(A)$ - marginal probability distribution of A
    \item $p(A\mid B)$ - conditional probability distribution of event A given B
\end{itemize}

\textit{Domain ($\mathscr{D}$)} -- can be defined as a combination of the input feature space $\mathcal{X}$ and an associated probability distribution $p(x,y)$, \ie $\mathscr{D}$ = \{$\mathcal{X}, p(x,y)$\}.
Here $p(x,y)$ represents the joint probability distribution over the feature-label space and can be decomposed into $p(x,y) = p(x)p(y\mid x)$ or $p(x,y) = p(y)p(x\mid y)$ , where $p(.)$ is the marginal distribution and  $p(.\mid.)$ is the conditional distribution.
The joint probability is used because the learning algorithms implicitly assume that each instance or sample $(x_i, y_i)$ is drawn from a joint distribution.

\textit{Dataset ($D$)} --
In the supervised setting, dataset $D$ consists of input-output pairs, \ie $D = \{(x_i,y_i)_{i=1}^n\}$, where $x_i$ is an m-dimensional feature vector, and $y_i$ is the response or output variable, which can be either a categorical variable or a real-valued scalar and $n$ is the number of labeled samples.

\textit{Task ($T$)} -- A loss function $L$ and a dataset $D$ can be used to define a task. \ie $T$ = \{$L$, $D$\}  \cite{meta_survey}.
There can be one task $T$ or a set of n tasks $\mathcal{T}$ = \{$T_1, T_2, ...,T_n$\} over a domain $\mathscr{D}$.

\subsection{Supervised Learning}\label{supervised_learning}

The objective of \ac{ML} algorithms is to learn from experiences.
Here, learning refers to improving by experience at a particular task  \cite{Mitchell_book}.
In supervised learning, one of the primary categories of \ac{ML} algorithms, these experiences happen to be labeled datasets, wherein the algorithm is fed with the inputs and the expected output, and the goal is to learn a mapping from the input to the output.
This article discusses only the various knowledge-transferring learning algorithms in the supervised sense. 

\textit{Supervised learning} -- for a specific domain $\mathscr{D}$ = \{$\mathcal{X}, p(x,y)$\}, the aim is to learn a predictive function $\mathcal{F_{\theta}}(x)$ from the training data \{$(x_i, y_i)_{i=1}^n$\}, where $x_i \in \mathcal{X}$ and $y_i \in \mathcal{Y}$ \ie the output label space and $\theta$ are the function parameters.
From a probabilistic point of view, $\mathcal{F_{\theta}}(x)$ can also be considered as the conditional probability distribution $p(y\mid x)$.
In the case of a classification task, when the response variable $y$ is categorical, \ie $y \in \{1,.....C\}$, and C is the number of categories, supervised learning aims to predict:
 \begin{equation}
 \centering
    \hat{y} = \mathcal{F_{\theta}}(x) = \argmax_{c\in[\![1,C]\!]}  ~~p(y=c \mid x)
\end{equation}   

This is known as Maximum A Posteriori (MAP), \ie the most probable class label. 

Defining the objective of supervised learning from the task point of view.
The loss function $L$ is a function of the model prediction, \ie $\hat{y}$ = $\mathcal{F_{\theta}}(x)$, and true labels y, \ie $L$ = $\sum_{i=1}^{N} f (\hat{y_{i}},y_i)$, for dataset $D =\{(x_1,y_1),..(x_N,y_N)\}$. 
In a supervised learning setting, for any task $T$, the objective is to optimize the parameters $\theta$ to minimize the loss. It can be expressed as --
\begin{equation}\label{eq2}
\centering
    \theta = \mathrm{arg} \min_\theta L(D;\theta)
\end{equation}
Based on the above explanation, many a time, as described by   \cite{9134370}, a task $T$ can also be defined as a combination of the label space $\mathcal{Y}$  and the predictive function $\mathcal{F_{\theta}}(x)$ \ie $T$ = \{$\mathcal{Y},\mathcal{F_{\theta}}(x)$\} = \{$\mathcal{Y}, p(y\mid x)$\}.
Therefore, for two tasks to be different, either the label space $\mathcal{Y}$ or the conditional probability distributions $p(y\mid x)$ must differ.

In conventional supervised learning, it is assumed to solve one task, \ie using one dataset over a domain, and so, the dataset is divided into development and test sets. 
The test set represents the future unseen data, but both sets are drawn from the same distribution and labeled. 
However, in reality, such a dataset is difficult to find, and therefore, the conventional single-task supervised learning algorithm fails to generalize.
For example, a model is trained using the handwritten digits dataset but is further used for inference on license plate images to detect the numbers automatically.
Therefore, for the model to perform on the data from another distribution, it must be trained again using the new data from the beginning. 
This sudden loss of previously learned information when information pertinent to a new task is introduced is commonly known as catastrophic forgetting  \cite{Hasselmo2017}.
This is one of the problems in traditional supervised learning and is said to be overcome in learning algorithms such as transfer learning, \ac{MTL}, meta-learning, and lifelong learning by sharing information between tasks.

\subsection{Transfer Learning} \label{sec_TL}

Most \ac{ML} models work under the assumption that the training and testing data are drawn from the same distribution (domain).
If the distribution is changed, the model must be trained again from scratch.
Transfer learning helps to overcome this issue.
Transfer learning refers to exploiting what has already been learned in one setting to improve learning in another  \cite{GoodBengCour16}.
Information transfer occurs from the source task (transferring knowledge to other tasks) to the target task (using knowledge from other tasks).
Such cases of sequential knowledge transfer are categorized as representational transfer approach  \cite{vilalta2017inductive}.
The domain of the source task and the target task may or may not be the same.

Assume only one source domain $\mathscr{D}_s$ = \{$\mathcal{X}_s, p_s$\} and target domain $\mathscr{D}_t = \lbrace\mathcal{X}_t, p_t\rbrace$ where $p_s$ and $p_t$ are the joint distributions of the source and target data, respectively. 
Let source data be $D_s$ = \{ $(x_{s_i},y_{s_i})_{i =1}^{n_s}$\}, where $x_{s_i} \in \mathcal{X}_s$ and $y_{s_i} \in \mathcal{Y}_s$ are the data instances and associated labels, respectively. 
Similarly, let the target data be $D_t$ = \{ $(x_{t_i},y_{t_i})_{i =1}^{n_t}$\}, where $x_{t_i} \in \mathcal{X}_t$ and $y_{t_i} \in \mathcal{Y}_t$. 
Usually, in transfer learning, it is considered that $0 \leq n_t << n_s$ \ie the source data is much larger than the target data.
Consider $T_s$  and $T_t$ to be the source and target tasks in their respective domains. 

Definition of transfer learning, as discussed by  \cite{Pan10asurvey}:\\
For a given source domain $\mathscr{D}_s$ and source task $T_s$, a target domain $\mathscr{D}_t$ and target task $T_t$, with $\mathscr{D}_s \neq \mathscr{D}_t$  or $T_s \neq T_t$, transfer learning aims to learn the target task $T_t$, \ie improve the learning of the target prediction function  $\mathcal{F_{\theta}}_t(x_t)$ in domain $\mathscr{D}_t$ using the knowledge gained by performing a source task $T_s$ in domain $\mathscr{D}_s$.
The objective in the source domain is to learn the source predictive function $\mathcal{F_{\theta}}_s(x_s)$ from the training data $D_s$.
Additionally, the dataset $D_s$ is not accessed during the transfer. 
\begin{figure*}[ht]
    \centering
    \includegraphics[width = 0.8\linewidth]{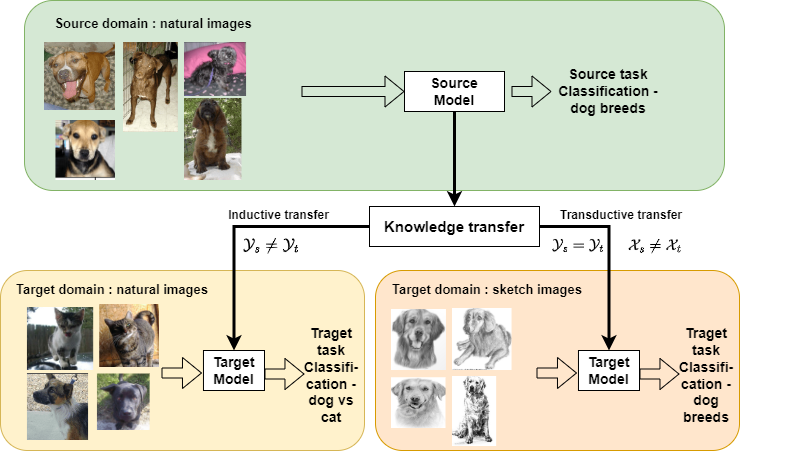}
    \caption{The figure shows two types of transfer learning from the source to the target task. The images are taken from the ImageNet dataset  \cite{5206848}.}
    \label{fig:TL}
\end{figure*}
In other words, it is solving target task $T_t$ after solving the source task $T_s$ by handing over the knowledge learned during task $T_s$.
In order to perform transfer learning, either of the two conditions in the definition should be satisfied. 
Since $\mathscr{D}$ = \{$\mathcal{X}$, p(X)\}, the condition $\mathscr{D}_s \neq \mathscr{D}_t$ suggests that either $\mathcal{X}_s \neq \mathcal{X}_t$ \ie the feature space of the domains are different, or $p_s(x) \neq p_t(x)$ which means even if the feature space are same, the marginal probability distribution of the source and target domain data are different.
Furthermore, as a task is defined as $\mathcal{T}$ = \{$\mathcal{Y}$, $p(y\mid x)$\}, the condition $\mathcal{T}_s \neq \mathcal{T}_t$ implies that either the label space of both the domains is different \ie $\mathcal{Y}_s \neq \mathcal{Y}_t$,  or the conditional probability distributions are different, \ie $p(y_s\mid x_s) \neq p(y_t\mid x_t)$
 
Based on these conditions, transfer learning can be classified as follows;

\begin{itemize}
   \item Inductive transfer learning - in this, $\mathcal{Y}_s \neq \mathcal{Y}_t$, and it does not matter whether the source and target input feature space are the same.
   Labeled data in the target domain are needed to learn a predictive model already trained in the source domain.
   Here, there are two possibilities depending on the labels;
   \begin{enumerate}
       \item \textit{The source domain has a lot of labeled data}; according to  \cite{Pan10asurvey}, this is considered as a limiting case of Multi-task learning  \cite{MTL-rich} (discussed in section \ref{sec_MTL)}). 
       But there is a difference: inductive transfer learning shares the knowledge of the source task to improve the learning of only the target task, while multi-task learning jointly learns both the task and attempts to improve the performance of both.  
       
       \item \textit{The source domain has no labeled data}; this converges to Self-taught learning  \cite{selftaught_raina}.
       In the case of unlabelled data, the source task learns a good feature representation and uses these learned feature representations to accomplish the target task. 
   \end{enumerate}
   
   \item Transductive transfer learning  \cite{TTL} - in this $\mathcal{Y}_s = \mathcal{Y}_t$ but  $\mathcal{X}_s \neq \mathcal{X}_t$, and it is assumed that the target domain has unlabelled data at the training time. 
   There are two cases- 
   \begin{enumerate}
       \item $\mathcal{X}_s \neq \mathcal{X}_t$ \ie feature space is different
       \item $\mathcal{X}_s = \mathcal{X}_t$, but $p_s(x) \neq p_t(x)$ \ie the marginal probability distribution is different for both domains. This case is related to domain adaptation  \cite{farahani2020brief}.
       Furthermore, covariance shift is a condition in domain adaptation when, along with $p_s(x) \neq p_t(x)$, the conditional distributions are constant $p_s(y\mid x) = p_t(y\mid x)$. 
       Data drift or concept shift is the case when $p_s(x) = p_t(x)$, while $p_s(y\mid x) \neq p_t(y\mid x)$.
       These are the types of domain shifts explained by  \cite{farahani2020brief}.
   \end{enumerate}
   
   \item Unsupervised transfer learning - there is no labeled data in either domain, and the aim is to solve unsupervised learning tasks in the target domain, for example, clustering, dimension reduction, etc., using a large amount of data in the source domain. 
   Self-taught clustering  \cite{dai2008self} is one such instance of unsupervised transfer learning.
\end{itemize}

There are two primary approaches for transferring information between source and target tasks  \cite{vilalta2017inductive}. 
The first is `literal transfer,' where the source model is used as it is for knowledge transfer.
Such an approach is usually referred to as feature extraction, which uses the source model architecture and model parameters to extract features from the target domain data to accomplish the target task.
Very often, only the higher (last) layers of the network are modified and trained, \ie the parameters are altered to adapt to the new data.
In contrast, the parameters of the lower (initial) layers are frozen, \ie the same as those from the source model.
The second is `non-literal transfer,' where the source model is modified before knowledge transfer.
This approach is usually referred to as fine-tuning, where the source model shares its parameters and architecture with the target task, but the source parameters only serve as initialization to the target network, and further training is required to adapt to the target domain data.

\subsubsection*{Example of transfer learning}
With limited training data, consider two target tasks, first classifying cats from dogs in the natural image domain and second identifying breeds of dogs using sketch images, as shown in Fig.~\ref{fig:TL}.
In addition, a source task identifies dog breeds with a larger dataset than the target task using natural images. 
Since conventional supervised machine learning models are data-hungry, it is difficult to accomplish the target tasks with less data, assuming that the source task with more data than the target performs sufficiently well. 
In this scenario, knowledge transfer from the source to the target, if performed effectively, may enhance the performance of the target task.
In the given example in Fig.~\ref{fig:TL}, two types of transfer are shown: inductive transfer where the domain is the same but the output labels are different ($\mathcal{Y}_s \neq \mathcal{Y}_t$) \ie the source task involves classifying breeds of dogs while the target task involves classifying cats versus dogs.
And the second is a transductive transfer, where the domains are different \ie natural images and sketch images, but the labels are the same ($\mathcal{Y}_s = \mathcal{Y}_t$) \ie both involve classifying breeds of dogs.

\subsection{Multi-task Learning} \label{sec_MTL)}
\begin{figure*}[ht]
    \centering
    \includegraphics[width = 0.85\linewidth]{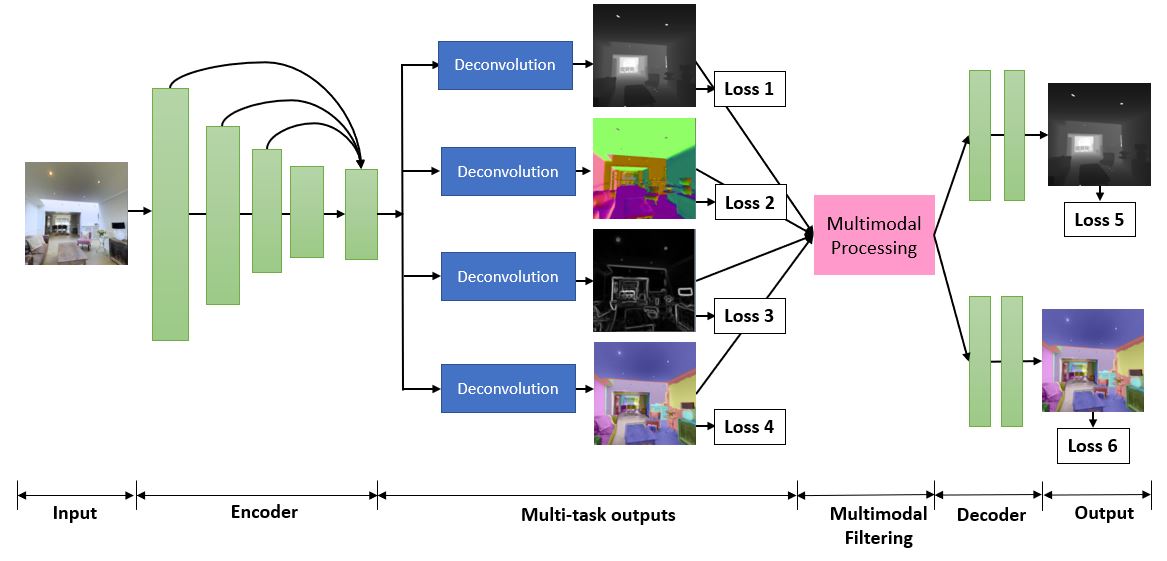}
    \caption{Illustration of PAD-Net architecture proposed by  \cite{Xu2018PADNetMG}, with four primary tasks: monocular depth estimation, surface normal estimation, edge detection, and semantic segmentation. Outputs are integrated to predict two output tasks: depth estimation and scene parsing. Here, Loss 1 - Loss 4 represent optimization losses for various tasks. All images in the figure are from the Taskonomy dataset  \cite{zamir2018taskonomy}.}
    \label{fig:PadNet}
\end{figure*}

\ac{MTL}, as explained by  \cite{MTL-rich}, is an inductive transfer approach that exploits the domain information in the training data of related tasks as an inductive bias to improve the generalization of all tasks. 
The underlying theory of \ac{MTL} is that the information gained while learning one task can help the other task learn better. 
In \ac{MTL}, all tasks are trained (or learned) together,  \cite{JMLR:v17:15-242} demonstrated that shared representations significantly improve the performance of the tasks compared to learning tasks individually.
Such an approach of concurrent knowledge transfer (parameter sharing) between tasks is known as the functional transfer approach  \cite{vilalta2017inductive}.
In conventional \ac{MTL} as discussed by  \cite{Crawshaw2020}, the parameter sharing approach is classified as -
  \begin{itemize}
    \item \textit{hard parameter sharing} - model weights (or parameters) are shared between multiple tasks, and each weight is modified to minimize multiple loss functions.
    This can be achieved as a result of \ac{MTL} architectures.
    \item \textit{soft parameter sharing} - tasks have separate weights, and the distance between the weights of the models for every task added to the joint loss function is minimized, similar to introducing a regularization term in the combined loss. Therefore, there is no explicit sharing of parameters; rather, models of different tasks are forced to have similar parameters. This is often introduced when there is a negative transfer between the task and the need to share less.
    The various optimization techniques help to achieve soft parameter sharing.     
\end{itemize}


Definition of \ac{MTL}, as discussed by  \cite{LL_book}: \\
Consider, $\mathcal{T}$ is an ensemble of N related but not identical tasks \ie $\mathcal{T}$ = \{$T_1,T_2....T_N$\} over a domain $\mathscr{D}$ and each task $T_i \in \mathcal{T}$ has training data $D^{tr}_{i}$ = $\{(x^{tr}_{i},y^{tr}_{i})_1, (x^{tr}_{i},y^{tr}_{i})_2,...,(x^{tr}_{i},y^{tr}_{i})_{m_{i}} \}$, where $m_i$ is the number of data instances for $i^{th}$ task. \ac{MTL} aims to jointly learn these multiple tasks \{$T_1, T_2....T_N$\} in order to maximize the performance of all the N tasks.
So, the objective of \ac{MTL} is to learn the optimal parameters $\theta^*$ in order to minimize the combined loss $L$ across each task. 
It can be expressed as:
\begin{equation}\label{eq:MTL}
\centering
    \theta^* = \min_{\theta\in \Theta=\cup_{i=1}^n\Theta_i} \sum_{i=1}^{N} L_i (\theta_i, D^{tr}_{i})
\end{equation}
Here, $L_i$ and $\Theta_i$ represent loss and parameters for $i^{th}$ task.
Equation~\ref{eq:MTL} exclusively demonstrates the scenario of hard parameter sharing.
The method for integrating the losses in this equation involves the straightforward approach of computing the total sum of the individual losses. 
Numerous other methodologies exist for merging losses in MTL. 
These include, but are not limited to, the incorporation of a weighted sum of losses that accounts for task significance, the utilization of uncertainty-based weighting of losses, and the implementation of gradient normalization. 
Further details on these techniques can be found in reference  \cite{vandenhende2021multi}.

As discussed in section \ref{negative_transfer}, in a multi-task setting, despite the tasks being related, negative transfer can exist.
This depends on the information sharing between the tasks and can be controlled by better \ac{MTL} architecture designs and task relationship learning.
In recent years, there has been significant research on creating shared architectures for \ac{MTL}.
 \cite{Crawshaw2020} surveyed common deep \ac{MTL} architectures used in computer vision, natural language processing, reinforcement learning, etc.
Deep \ac{MTL} architectures can be divided into two types of modules  \cite{GoodBengCour16}:
\begin{itemize}  
    \item[] \textbf{Generic modules:} These are shared across all tasks, and the parameters benefit from  the data of all tasks (corresponding to $\Theta_g = \bigcap_{i}\Theta_i$);
    \item[] \textbf{Task-specific modules:} These are dedicated modules for each task, and the parameters benefit from the instances of the particular task (corresponding to $\Theta_i - \Theta_g$).
\end{itemize}

Note that in this context, modules are combinations of layers of neural networks or the \ac{CNN}.
As a result of these modules, the parameters are divided into shared parameters and task-specific parameters.
The best performance of \ac{MTL} models is achieved only when there is balanced sharing because too much sharing can cause negative transfer, and too little sharing can inhibit the effective leveraging of information between tasks. 
Therefore, to create an effective \ac{MTL} architecture, it is important to analyze how to combine the shared modules (layers) and task-specific modules and what portion of the model's parameters is shared between tasks.

\subsubsection*{Example of MTL}
Consider a dataset with images of natural scenes.
Every image is labeled for a number of tasks, such as surface normal estimation, semantic segmentation, edge detection, and depth estimation. 
For conventional machine learning, all four are different tasks, and it is required to train different models for each of the problems. 
\ac{MTL}, however, exploits the fact that all the tasks are related and use the same input image.
Therefore, they can be trained together in an \ac{MTL} architecture so that they share the representations between tasks and encourage the model to generalize better than single-task learning.
A similar architecture was proposed by  \cite{Xu2018PADNetMG} in Fig .\ref{fig:PadNet}.
Furthermore, in the above example, additional tasks such as learning image compression and decompression, colorization of grayscale images, denoising of images, etc., can also be integrated into the architecture.

\subsection{meta-learning} \label{sec_meta}
\begin{figure*}[ht]
    \centering
    \includegraphics[width = 0.85\textwidth]{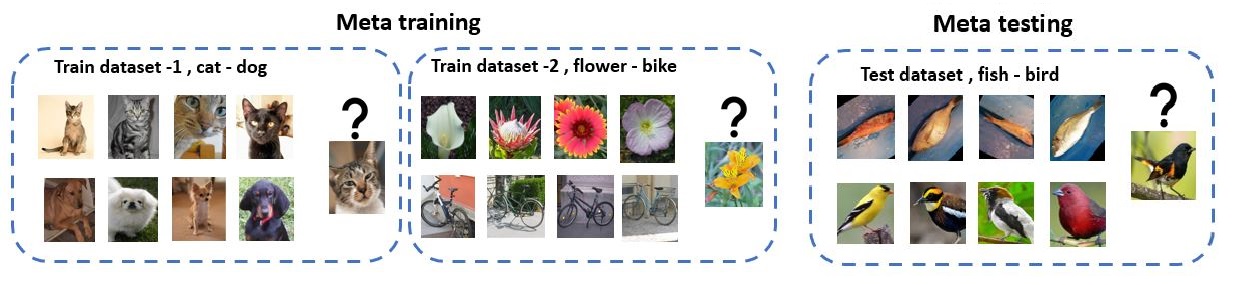}
    \caption{An example of meta-learning illustrating 4 shot 2 class image classification.}
    \label{fig:meta_learning}
\end{figure*}

meta-learning better known as ``\textit{Learning to learn}''  \cite{Thrun98, baxter1998theoretical}, is a learning paradigm that aims to improve the learning of new tasks with lesser data and computation by exploiting the experience gained over multiple training episodes for various tasks. The conventional \ac{ML} algorithm employs multiple data instances for better model predictions, while meta-learning uses multiple learning instances to improve the performance of a learning algorithm.

meta-learning can be very broadly defined as:\\
Assuming a set of M source tasks $\mathcal{T}_s$ = $\{T_{s_{1}}, T_{s_{2}},...,T_{s_{M}}\}$, sampled from a distribution $p(\mathcal{T})$,  with source datasets $D_{s}$. 
And Q target tasks $\mathcal{T}_t$ = $\{T_{t_{1}}, T_{t_{2}},...,T_{t_{Q}}\}$ with target datasets $D_{t}$.
meta-learning is to train a model on the M source tasks using data $D_{s}$, such that it generalizes well on a new unseen target task, which leads to:
\begin{itemize}
\itemsep 0em 
    \item Computational efficiency - faster training of the target task using data $D_t^{train}$ 
    \item Data efficiency - good training with fewer target data instances 
    \item Effective knowledge transfer - good performance on the target test data $D_t^{test}$ 
\end{itemize}

Note - The source and target tasks in conventional meta-learning are in the same domain  \cite{pmlr-v70-finn17a}, though there are some works that focus on domain generalization in which the domains for the source and target tasks and domains can be different such as  \cite{li2018learning, qiu2021meta, triantafillou2019meta}.

Let a task $T$ be defined as, $T = \{L, D\}$, where $D =\{(x_1,y_1),..(x_N,y_N)\}$ is training dataset and $L$ is the loss function.
For a single task conventional supervised \ac{ML} algorithm, the aim of learning a model $\hat{y} = f_\theta(x)$ parameterized by $\theta$, is accomplished by solving (same as Eq. \ref{eq2}):
\begin{equation}
\centering
    \theta^*(T) = \mathrm{arg} \min_\theta L(D;\theta,\phi)
\end{equation}
Here, $\phi$ denotes the ``how to learn'' assumptions  \cite{meta_survey}, for example, the optimizer for $\theta$, choice of hyper-parameters, etc.  
A pre-specified $\phi$ can help to achieve significant performance as compared to the case when it is absent.

meta-learning involves learning a generic algorithm by training over several tasks that enable each new task to learn better than the previous.
Therefore, for a distribution of task $p(\mathcal{T})$, meta-learning becomes:
\begin{equation}\label{meta}
   \min_\phi \displaystyle \mathop{\mathbb{E}}_{T \sim p(\mathcal{T})} L (D;\theta^{\star}(T), \phi)
\end{equation}
where $L(D;\theta^{\star}(T),\phi)$ evaluates the model's performance trained using $\phi$ on task $T$. 
$\theta^{\star}(T)$ is the optimal parameter learnt for task $T$.
Here, the parameter $\phi$ is the \textit{meta knowledge} or across task knowledge  \cite{meta_survey}.

To solve the meta-learning problem, assuming M source tasks $\mathcal{T}_s$ sampled from $p(\mathcal{T})$, having dataset $D_{s} = \{(D_s^{train}, D_s^{val})^{(1)},...,(D_s^{train}, D_s^{val})^{(M)}\}$, with train (support) and validation (query) sets. 
Also, Q target tasks with data $D_{t} = \{(D_t^{train},D_t^{test})^{(1)},...,(D_t^{train},D_t^{test})^{(Q)}\}$, \ie each task with train and test set.
In the meta-learning taxonomy, these tasks (both source and target tasks) are frequently referred to as learning episodes  \cite{meta_survey}.

The meta-learning objective in eq.[\ref{meta}] is obtained in two stages, 
\begin{itemize}
    \item[] \textbf{Meta training} - The meta training stage can be posed as a bi-level optimization problem, where one optimization contains another optimization as a constraint.
    Here, an inner learning algorithm solves a task, defined by dataset $D_s^{train(i)}$ and objective function $L^{task}$. 
    While in meta training, an outer (meta) algorithm updates the inner algorithm in order to improve the outer objective $L^{meta}$. 
    So, meta training can be formulated as:\\
    (outer Objective)
    \begin{equation}\label{phi}
        \phi^* = \mathrm{arg}\min_\phi \mathop{\mathbb{E}}_{\mathcal{T}_s \sim p(\mathcal{T})}  L^{meta}~(\theta^{*(i)}(\phi), \phi, D_s^{val(i)})
    \end{equation}
    where,\\
    (inner objective)
    \begin{equation}\label{theta}
        \theta^{*(i)}(\phi) = \mathrm{arg}\min_\theta ~ L^{task}~(\theta, \phi, D_s^{train(i)})
    \end{equation}

    $\phi^*$ has all the information of source tasks (or data) to solve new tasks.
    So, the inner objective corresponds to task-specific learning, while the outer objective corresponds to multiple-task learning. 
    \item[] \textbf{Meta testing} - The meta testing stage is often referred as \textit{adaptation} stage.
    This stage uses the meta knowledge or meta parameters ($\phi^*$) to train the model on unseen target tasks. For an $\mathrm{i^{th}}$ target task, meta testing involves training on the $D_t^{train(i)}$ to minimize the loss $L^{test}$ and evaluating the performance on $D_t^{test(i)}$, for optimal parameter $\theta^{*(i)}$ given by:
    \begin{equation}
     \theta^{*(i)} = \mathrm{arg}\min_\theta~ L^{test}(\theta,\phi_{\star},D_t^{train(i)})
    \end{equation}
\end{itemize}

However, \ac{MTL} can conceptually be seen as a special case of meta-learning if $\theta = \phi$ in the meta-training phase, as there will be only one optimization objective and multiple tasks for training. 
Despite this similarity, there are many differences that persist between meta-learning and \ac{MTL}, which are discussed in Sec. \ref{comparisons}.

\subsubsection*{Example of meta-learning}

Fig. \ref{fig:meta_learning} shows an example of meta-learning, where the source tasks are: Task 1-  classification between cats and dogs, and Task 2- classifying flowers from bikes. 
The target task is to classify between images of fish and birds that the model has not seen during the meta-training phase. 
This is a classic example of 4-shot 2-class image classification, where the objective is to learn to identify the categories only by 4 images, and every task has 2 classes or labels. 
So, meta-learning enables the model to learn fast in a few instances of fish and bird images during meta-testing by utilizing the meta-knowledge gained from meta-training of the source tasks.
The number of source tasks can be increased in order to achieve better generalization.

\subsection{Comparisons} \label{comparisons}
\begin{figure}[ht]
    \centering
    \includegraphics[width = 0.98\linewidth]{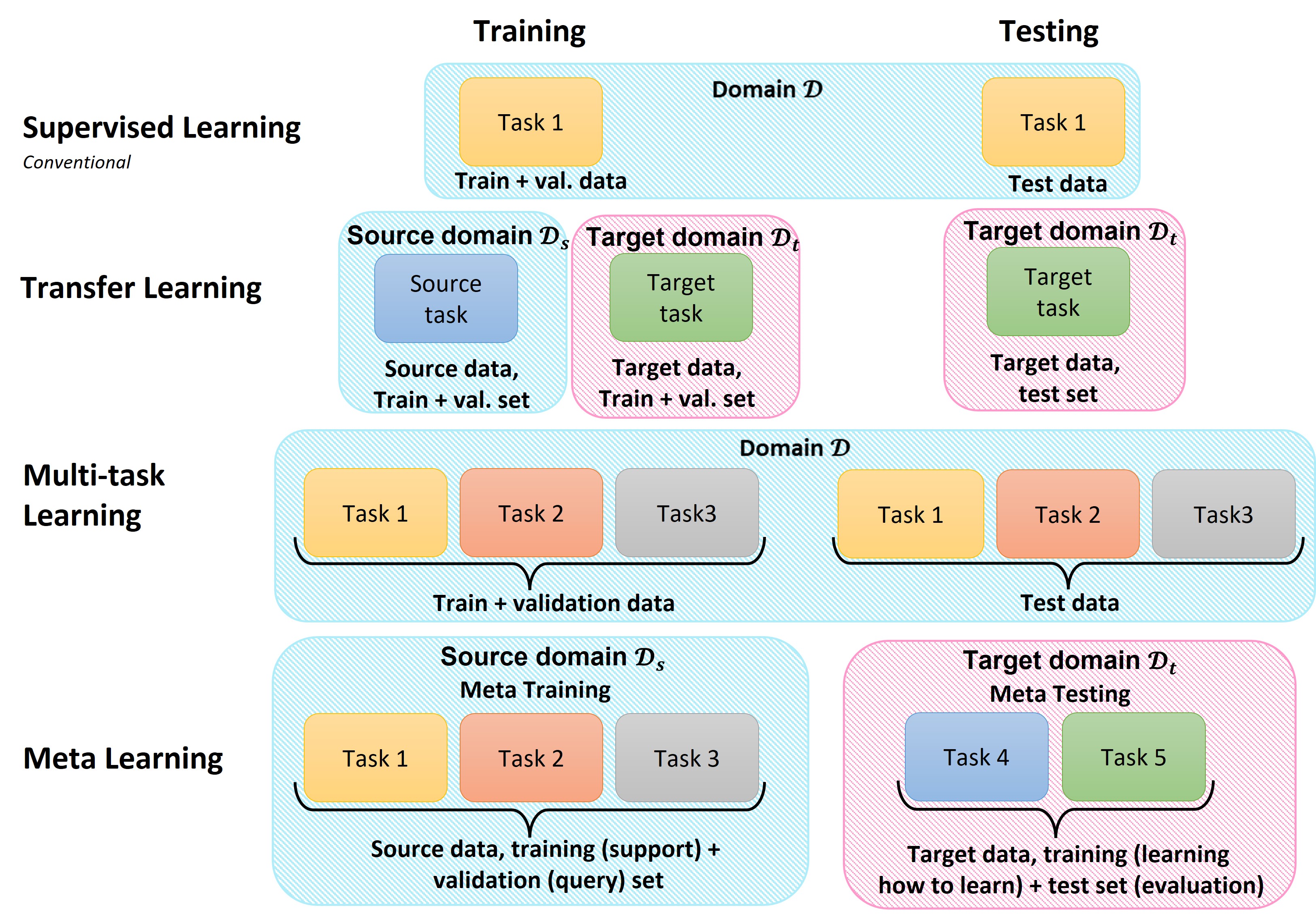}
    \caption{A comparative representation of how tasks are introduced in the learning paradigms}
    \label{fig:comparision}
\end{figure}

All three learning paradigms, as described in Sec.~\ref{learning_paradigms}, communicate knowledge between tasks, but the differences reside in the specifics of when, how, and what is shared.
As an additional distinction, the way in which tasks are introduced varies throughout the various learning paradigms.
Fig.~\ref{fig:comparision} depicts a summary of the comparison of these learning paradigms on the basis of the tasks and their domains throughout the training and testing stages.
It is important to highlight that in Fig.~\ref{fig:comparision}, the source and target domains in transfer learning and meta-learning are believed to be distinct in order to simplify the process of comprehension.
According to the definitions in Sec.~\ref{learning_paradigms}, domains can be similar, but in that case, the source and target tasks must be distinct.
Consequently, Fig~\ref{fig:comparision} illustrates the case when the source and target domains and tasks are both distinguishable. 
This section presents a comparative view of the \ac{MTL}, meta-learning, and transfer learning.
It might not cover all the differences, but it does focus on significant dissimilarities and similarities.
\begin{table*}[ht]
  \centering
    \caption{Comparison between transfer learning, \ac{MTL}, and meta-learning (gradient-based meta-learning algorithms, \eg MAML  \cite{pmlr-v70-finn17a}).}
    \begin{tabular}{p{0.2\linewidth} p{0.2\linewidth} p{0.2\linewidth}  p{0.2\linewidth}}
    \hline
      \textbf{Features} &\textbf{Transfer Learning} & \textbf{MTL} & \textbf{meta-learning} \\ \hline \hline
    What is shared? & model architecture, parameters & parameters and features & parameters, hyper-parameters, architectures, and many more\\
    When is it shared? & after training source tasks & while training the tasks& after training the source tasks\\  
    Why is it shared? & data and label efficient & shared representation improve all tasks & quickly learn new task\\  
    How is it shared? &from source to target task&between all task while training&from source to target task\\
    Type of knowledge transfer & representational transfer & functional transfer & representational transfer\\
    Heterogeneous tasks & tasks need not be identical & tasks need not be identical & tasks should match. \\
    Training data requirement & source dataset $>>$ target data & large training data required & few-shot variant : less training data required\\
    Computation efficiency & most efficient (comparatively) & more efficient (comparatively),\hspace{5em}more tasks more computation  & meta training - not-efficient\\
    Structure/technique & mini-batch & mini-batch & episodic learning\\
    of training & one batch one task & one batch multiple task & one episode one task\\
    Two-level optimization &$\times$ & $\times$ &\checkmark \\
    Negative transfer &\checkmark & \checkmark & $\times$\\
    Multi-modal inputs & \checkmark & \checkmark & \checkmark \\     
     \hline
    \end{tabular}
    
    \label{tab:proscons}
\end{table*}


\subsubsection{Transfer learning and  MTL}
The tasks involved in transfer learning and \ac{MTL} can be heterogeneous, \ie the nature of the tasks can be different, such as classification, segmentation, and regression. 
These learning paradigms can share features and parameters between tasks.
Because of the inductive transfer approach, \ac{MTL} is also considered a type of transfer learning  \cite{Pan10asurvey}, but they are very different.
\ac{MTL} learns many tasks together, while this is not the case in transfer learning, where the source task is trained, and the information is transferred for learning the target task. 
In other words, transfer learning trains tasks sequentially, meaning one follows the other, resulting in information transfer occurring after the source tasks are trained. In contrast, Multi-Task Learning (MTL) typically involves training tasks simultaneously or jointly, allowing for information transfer to take place among all tasks during training. 
In \ac{MTL}, the goal is to generalize the performance of all tasks, while in transfer learning, the focus is only on the generalization of the target domain.
Transfer learning is a logical explanation for multi-task learning but not vice versa.

\subsubsection{Meta-learning and Transfer learning}
In both of these learning paradigms, the target tasks are trained after the successful training of the source task, \ie sequentially. 
Both algorithms aim to achieve better generalization on the target task.
The key difference between them lies in the training technique for the source tasks.
In transfer learning, there is no meta-objective when deriving priors (parameters) from learning the source task. 
In contrast, in meta-learning, the priors are extracted as a result of the outer-loop (or meta) optimization, and these are evaluated while learning a new task.
Only model parameters are shared in transfer learning, while meta-learning transfers a variety of meta representations. 
In few-shot learning applications of meta-learning (like gradient-based meta-learning \eg MAML), the meta-training (source) and meta-testing (target) conditions must match, \eg if the source tasks are binary classification problems, then the target task necessarily should be binary classification.
In contrast, transfer learning is possible on diverse tasks.
Fundamentally, meta-learning can be viewed as a mechanism to facilitate transfer learning, given that information is conveyed from source tasks to target tasks. However, the nature of the information transferred and the methodology employed for its collection diverge substantially from traditional transfer learning. 

\subsubsection{Meta-learning and Multi-task learning}
Similar to transfer learning, \ac{MTL} has single-level optimization, \ie no meta objective.
\ac{MTL} aims to solve a fixed number of known tasks jointly, whereas meta-learning addresses solving unseen tasks.
meta-learning, as discussed earlier, only works with homogeneous tasks during meta training and testing.
There is a great deal of research in the field of multi-modal meta-learning  \cite{ma2022multimodality}, but it is distinct from the application of meta-learning to non-homogeneous tasks, as the former focuses on diverse inputs and the latter on diverse tasks.
On the other hand, \ac{MTL} can handle heterogeneous tasks and also a variety of inputs. 
The source tasks are trained sequentially for many iterations during meta-training in meta-learning. 
A new unseen task is learned at the time of meta-testing; in contrast, in MTL, the training and testing are jointly performed for all tasks. 
Therefore, the knowledge is shared between the tasks at the time of training in \ac{MTL}, but for meta-learning, the model saves the prior knowledge from the source tasks to be used during meta testing the target tasks. 
Although the meta-update (outer loop) in meta-learning can be considered a form of multi-task learning, a combined loss from multiple tasks is used for the weight update of the meta-model during the meta-training phase.

\section{Ensemble of the learning paradigms} \label{ensemble}
The strengths and weaknesses of the learning paradigms detailed in Sec. \ref{learning_paradigms} are summarized in Table \ref{tab:proscons}.
Each algorithm has some drawbacks, which are overcome by another learning algorithm. 
For example, meta-learning supports homogeneous tasks only, while \ac{MTL} may integrate heterogeneous tasks and multimodal inputs. 
Introducing a new task is much simpler in meta-learning and transfer learning than \ac{MTL}.
For these reasons, these algorithms are coupled together in the literature to utilize their best features.
This section discusses the research performed in various ensembles of these algorithms.

\subsection{Research in Multi-task meta-learning}
Insights from both meta-learning and multitask learning can be fused to achieve the best of both worlds, \ie efficient training of multiple heterogeneous tasks, a feature of \ac{MTL}, and quickly adapting new tasks, a  characteristic of meta-learning.
Thus providing a quick, effective, and adaptable learning mechanism.
Numerous studies have demonstrated the effectiveness of combining both of these learning methods.
The articles that integrate \ac{MTL} and meta-learning can be broadly categorized into two types based on the taxonomy used for the ensemble.
\begin{itemize}

    \item[] Meta Multi-task Learning: It employs meta-learning to acquire meta-knowledge (\eg task relationships) to transfer it to a new task to improve \ac{MTL}.
    \item[] Multi-task meta-learning: In order to improve meta-learning, it introduces multi-task learning, which facilitates training with diverse tasks and efficient training processes, which lead to enhanced feature learning.
\end{itemize}
In general, the taxonomy utilized to combine the two algorithms has no logical justification.
As long as the main objective is to leverage the properties of meta-learning and \ac{MTL}, they are believed to be used interchangeably to integrate these learning mechanisms.
Based on the above categories, a detailed review of the related research work in various application domains is as follows:

\subsection*{Meta Multi-task Learning} 

The article  \cite{chen2018meta} suggested a function level information sharing method for \ac{MTL}, with a basic \ac{LSTM}  \cite{hochreiter1997long} that is task-specific and a meta \ac{LSTM} that is shared across all tasks.
This work focused on the two \ac{NLP} tasks of sequence tagging and text classification. 
A shared meta-network is used in this work to learn the semantic composition among the different tasks that act as meta-knowledge.
The semantic composition functions play a vital role in neural representation learning of text sequences; therefore, this work focused on learning generalized functions across tasks that can be transferred to new tasks. 
The proposed design is a multi-task architecture using a two-level meta-learning optimization technique.
They demonstrated that the proposed network performs significantly better than using \ac{MTL} and single-task learning on several datasets.
However, they do not present the performance when a new task is introduced.

\begin{table*}[p]
\centering
\caption{A compilation of the various types of tasks and datasets used by the works detailed in this article.}
\label{tab:datasets}
\begin{tabular}{p{0.06\linewidth}  p{0.3\linewidth}  p{0.3\linewidth} p{0.24\linewidth}}
\hline
\textbf{Articles} & \textbf{Tasks} & \textbf{Datasets} & \textbf{Comments} \\
\hline
\hline 
\multicolumn{4}{c}{MULTI-TASK LEARNING + META-LEARNING }\\
\hline
\hline 
\multirow{4}{*}{ \cite{chen2018meta}} & text classification & amazon product reviews  \cite{14classificationdataset} & each domain is a task\\
& sequence tagging & Wall Street Journal (WSJ) \cite{WSJ_PTB} & \\
& chunking & CoNLL 2000  \cite{2000conll} & \\
& Named Entity Recognition & CoNLL 2003  \cite{conll2003} & \\
\hline 
\multirow{3}{*}{ \cite{liu2018meta}} & test classification & combination of many datasets  \cite{TextClassification} & product and movie review datasets \\
 & sequence labelling &  CoNLL 2000 \& 2003  \cite{2000conll}  \cite{conll2003} & \\
 & image aesthetic assessment & multi-domain AVA dataset  \cite{ImgAestheticAssessment} & \\
 \hline

\multirow{4}{*}{ \cite{tarunesh2021meta}} & classification, structured prediction, & XTREME multilingual  \cite{XTREME}&  \\
& question answering, retrieval & & \\
 & part of speech & Universal Dependency v2.5 treebank  \cite{nivre-etal-2020-universal} & \\
  & named entity recognition &  WikiAnn  \cite{pan-etal-2017-cross} & \\
\hline
\multirow{2}{*}{ \cite{chen2020st2}} & text style transfer & literature translation dataset  \cite{trans_data} &\\
& & grouped standard datasets  \cite{GroupedStandardDataset} & combined Yelp/Amazon review, GYAFC datasets and more\\
\hline
 \cite{lee2021generating} & personalized dialogue generation & Personal Chat dialogue  \cite{PersonalChat} & each persona is a task\\
\hline
\multirow{2}{*}{ \cite{lekkala2020attentive}} & scene classification, depth estimation,  &  Multi Meta Tasks dataset (MMT) & dataset of datasets  \cite{8099744,Silberman:ECCV12,diode_dataset} \\
& surface normal estimation, vanishing point & &  \cite{NIPS2016_90e13578,7968387}\\
\hline 

\multirow{2}{*}{ \cite{cai2020meta}} & speech emotion recognition & IEMOCAP  \cite{IEMOCAP} & introduce auxiliary tasks of \\
 & & & valence, activation \& dominance classification\\
\hline
 \cite{Liu2020MultitaskLV} & mortality prediction of rare disease & EHR  \cite{EHR}, MIMIC-III  \cite{MIMIC-III}, eICU   \cite{EHR} & \\
\hline
\multirow{2}{*}{ \cite{upadhyay2023multitask}} & semantic segmentation, depth estimation, & tiny-taskonomy  \cite{zamir2018taskonomy}, NYU-v2  \cite{Silberman:ECCV12} & \\
& surface normal estimation, edge detection & & \\
\hline
\multirow{2}{*}{ \cite{Yang_2022}} & graph classification, node classification & parkinson's progression markers initiative  \cite{MAREK2011629}& \\
& link prediction & HIV and bipolar disorder dataset &  (not publicly available)\\

\hline
\hline 
\multicolumn{4}{c}{MULTI-TASK LEARNING + TRANSFER LEARNING }\\
\hline
\hline 
\multirow{3}{*}{ \cite{Ye_2018}} & generate molecule structure (source task) & large bioactivity data set  \cite{C7SC02664A} & \\
 & predict human pharmacokinetic parameters & pharmacokinetic dataset  \cite{drugbank} & \\
  &(target task) & & \\
\hline
\multirow{3}{*}{ \cite{s20247205}} & multi-task bearing fault diagnosis &  Case Western Reserve University data  \cite{cwru2017} & different datasets generated using \\
 & speed detection, health detection & & a induction motor (2 hp),  \\
 & (for both source and target domains) & & a dynamometer, and a transducer\\
 \hline
\multirow{2}{*}{ \cite{cruz-etal-2020-localization}} & fake news classification & fake news Fililino  \cite{WikiText_Filipino} & pretrained ULMFit  \cite{2018arXiv180106146H} and BERT\\
& language modelling & &  \cite{devlin2019bert} are used as source models \\
\hline 
\multirow{2}{*}{ \cite{dong_0216046}} & medical name entity recognition & electronic medical record datasets & dataset is Chinese\\
 & parts-of-speech tagging &  \cite{EMR_A1}  \cite{EMR_A2} & \\
\hline 
\multirow{2}{*}{ \cite{taslimipoor_cross}} & identification of multiword expression (MWEs) & multilingual dataset from the PARSEME & different languages contribute to \\
 & dependency parsing  & project  \cite{PARSEME} & source and target domains\\
\hline
\multirow{2}{*}{ \cite{8264705}} & neural decoding, image reconstruction & fMRI datasets:  Vim-1  \cite{kay2008}, FaceBold \cite{du2022} & \\
 & & image datasets: ImageNet-1K  \cite{5206848}, CelebA  \cite{liu2015deep} & \\
 \hline
\cite{Nguyen_2017} & album classification, (comic) character analysis & Manga109  \cite{9069265} & multi-modal : images and text\\
\hline
\multirow{2}{*}{ \cite{8920992}} & hyper-spectral image classification (target task) & Pavia university scene dataset  \cite{pavia_dataset}& similarity learning between same obj-\\
 & & & ects in different domains (source task)  \\
\hline
\multirow{2}{*}{ \cite{Dong2021}} & interaction patterns between particular virus and human protein & APID  \cite{APID}, IntAct  \cite{IntAct}, VirusMentha  \cite{VirusMentha}, UniProt  \cite{UniProt}, DeNovo SLIM  \cite{DeNovo} , VirusMINT  \cite{VirusMINT}, HPIDB  \cite{HPIDB} & \\ 
\hline
\multirow{1}{*}{ \cite{xiao2024integrated}} & bioaccumulation parameters prediction & KOW, FBM, FBA, FBC  \cite{xiao2024integrated}& predicting multiple parameters (MTL)\\ 

\hline
\hline 
\multicolumn{4}{c}{META-LEARNING + TRANSFER LEARNING}\\
\hline
\hline 
 \cite{Sun_2019_CVPR} & few-shot classification & miniImageNet  \cite{MiniImagenet}, Fewshot-CIFAR100  \cite{FC100} & \\
\hline
 \multirow{2}{*}{ \cite{Soh_2020_CVPR}} & Image super-resolution & DIV2K  \cite{8014884} & various degradations make several \\
& & Set5  \cite{BMVC.26.135}, BSD100  \cite{937655}, Urban100  \cite{7299156} & learning episodes or tasks\\

\hline
\multirow{2}{*}{ \cite{Willard_2021}} & depth specific temperature prediction & North American Lake Data  Assimilation  & transfer learning from monitored \\
& & System  (NLDAS-2)  \cite{mitchell_lake}, U.S. geological survey's science-base platform  \cite{read_lake} & lakes to un-monitored lakes \\
\hline
\multirow{3}{*}{ \cite{Song_UAV}} & vehicle tracking in UAV videos & VisDrone2018  \cite{zhu2018vision}, UAV123  \cite{UAV123}, & transfer learning across landscapes \\
& &  Drone Tracking Benchmark (DTB)  \cite{Li_Yeung_2017}& \\
& &  large-scale UAV-Vehicle  \cite{du2018unmanned} & \\
\hline
\multirow{3}{*}{ \cite{10506110}} & hyperspectral image super-resolution & CAVE  \cite{yasuma2010generalized}, Harvard  \cite{chakrabarti2011statistics}, & transfer learning from natural to \\
& &  Pavia  \cite{pavia_dataset}, Chikusei  \cite{yokoya2016airborne} & remote sensing, meta-learning across \\
& &  Cuprite  \cite{cupriteDataset} & hyperspectral bands\\
\hline
\multirow{3}{*}{ \cite{She_2024}} &  gearbox fault diagnosis & planetary gearbox dataset  \cite{She_2024} & transfer learning from various   \\
& &  & operating conditions, meta-learning  \\
& &  & across different fault types\\
\hline
\multirow{3}{*}{ \cite{wu-etal-2023-good}} & Named Entity Recognition, & Wikipedia Multilingual Corpus  \cite{devlin2019bert} & transfer learning from high to low\\
& Machine Reading Comprehension & WikiAnn  \cite{pan-etal-2017-cross}, TydiQA  \cite{clark2020tydi}  & resource languages, meta-learning \\
&  &  & many low-resource languages\\
\hline
\multirow{2}{*}{ \cite{water_temp}} & water temperature forecasting & Delaware river dataset  \cite{usgs_nwis, water_quality_portal, regan2018} & transfer learning from historical data,\\
 &&& meta-learning across temp. periods\\
\hline
\end{tabular}
\end{table*}

The article  \cite{liu2018meta} presented a communication strategy between the tasks in the multi-task setting by introducing the Parameters Read-Write Networks (PRaWNs). 
The authors argue that different tasks may require different communication strategies between the model's components in the context of NLP. 
The proposed approach addresses this challenge by introducing a meta-learner that learns to adapt to the communication strategy required by each task.
By communication strategy, they mean explicitly allowing tasks to pass gradients rather than constantly updating the parameters. 
Consequently, gradients are transferred pairwise and list-wise between tasks, with list-wise gradients accounting for task relatedness.
In \ac{MTL}, the shared feature space is usually entangled since tasks share information; allegedly, the approach proposed in this work enabled the separation of the features of various tasks.
The research examines the proposed network for text classification, sequence tagging, and image aesthetic evaluation tasks in both in-task (same dataset) and out-of-task (new dataset) contexts.
The results show that the proposed approach outperforms existing methods, achieving state-of-the-art results even in the out-of-task setting.

The article  \cite{tarunesh2021meta}  proposed a novel approach for multi-task and multilingual learning that leverages the concept of meta-learning.
This article assesses the multiple tasks of question answering, parts of speech tagging, name entity recognition, paraphrase identification, and natural language inference for various languages. 
The authors argue that existing multi-task and multilingual learning methods suffer from several limitations, including the need for extensive training data, difficulty in handling diverse tasks and languages, and limited ability to adapt to new tasks and languages quickly. 
The proposed approach aims to address these limitations by using meta-learning to learn a shared representation of the input data that can be used to solve multiple related tasks and languages.
`Reptile'  \cite{reptile}, a first-order meta-learning algorithm, is used on episodes of task-language pairs.
In addition, they propose integrating sampling approaches such as heuristic and parametric sampling into meta-learning to enhance the performance of the model.
This method can be highly beneficial for languages with minimal data resources, as it performs exceptionally well with zero-shot new target languages.

The article  \cite{cai2020meta} presented \ac{MTL} and meta-learning for speech emotion recognition. 
Although the main focus of this work is on speech emotion recognition, they also incorporated a few auxiliary classification tasks such as valence (positive or negative), activation (calm or excited), and dominance (passive or aggressive).
Identifying the relationship between these factors (auxiliary tasks) and human emotions (main task) is the main challenge in speech emotion recognition.
A multi-task model in this work undergoes two phases of training -- the multi-train stage and the knowledge transfer stage. 
In the multi-train stage, all the auxiliary tasks are trained together, and the meta-information is then shared in the knowledge transfer stage for the emotion recognition task. 
\ac{MAML}, a gradient-based meta-learning technique, is employed for accumulating the meta-knowledge during the multi-train stage.
This approach, therefore, leveraged the similarities between related tasks to improve the performance of speech emotion recognition. 

The work  \cite{Liu2020MultitaskLV} addressed the problem of data scarcity and disease diversity in the prediction of mortality in rare diseases by combining a multi-task architecture with the meta-learning optimization method \ac{MAML}  \cite{pmlr-v70-finn17a}.
The many tasks, in this case, are recognizing the temporal occurrences of rare diseases, and the input is multi-modal, including text, photos, signals, and so on.
This paper introduces the Ada-SiT (Adaptation to Similar Task) learning method, in which task similarity is assessed during meta-training and utilized to share initialization for faster adaption of new tasks.
The proposed method uses a two-stage \ac{MTL} framework. The first stage trains a base model on a set of related diseases, and the second stage adapts the base model to the target rare disease using limited data.
The authors evaluate the proposed approach on a dataset of 89 rare diseases and demonstrate that the proposed method outperforms several baseline methods for mortality prediction. The results show that the adaptation to similar tasks in the second stage of the MTL framework improves the performance of mortality prediction on rare diseases with limited data.

\subsection*{Multi-task meta-learning}

Using meta-learning and \ac{MTL},  \cite{lee2021generating} improved the generation of personalized dialogue by attempting to circumvent the issue of an extensive dataset for each individual and integrating pre-defined persona information.
The proposed method consists of three main stages. First, a large corpus of dialogue data is collected, and user profiles are created based on demographic information and user interactions.
Second, a multi-task network is trained using meta-optimization to generate responses tailored to each user's profile.
Finally, the trained network generates real-time personalized responses during conversations with users.
This study includes the persona reconstruction task only during the meta-training phase as an auxiliary task to collect persona information as meta-knowledge for the dialogue-generating task.
Thus, the model can generate dialogues for new users during meta-testing.
Multi-Task meta-learning (MTML), which incorporates losses from both tasks, and Alternating Multi-Task meta-learning (AMTML), which acts alternately on the tasks of generation and persona reconstruction, are introduced.
Overall, the article presents a promising approach to generating personalized dialogue that can improve user engagement and satisfaction in various applications, such as chatbots, virtual assistants, and customer service agents.

A very similar work  \cite{chen2020st2} that aimed at text style transfer with limited data, essentially paraphrasing text from one writing style to another.
The problem of text style transfer involves changing the style of a piece of text while preserving its content.
The authors used a \ac{MTL} architecture to learn a shared representation of the language that can capture both the content and style of the text. The shared representation is learned by training the network on a set of related tasks, such as sentiment analysis, named entity recognition, and part-of-speech tagging. The goal is to learn a representation that can capture the underlying structure of the language and the relationship between content and style, which can be used to perform text style transfer.
It uses the \ac{MAML}  \cite{pmlr-v70-finn17a} approach for few-shot text style transfer, where an episode corresponds to a pair of styles for multiple tasks.
Because of the meta-learning framework, this permits the transfer (to and from) writing styles with limited training data and data that the model has never seen before.
In text style transfer, the proposed methodology outperforms the state-of-the-art method.

 \cite{lekkala2020attentive} proposed a novel approach to multi-task meta-learning that improves the efficiency and performance of neural networks by allowing them to reuse previously learned features. 
The approach uses a new module called Attentive Feature Reuse (AFR) that selectively reuses features from past tasks based on their relevance to the current task.
The network in this work has three parts: the shared backbone network layers, the task-specific layers or the task heads, and the AFR module.
The parameters of the shared backbone network are weighted according to the importance of a specific task.
At the same time, the task-specific layers are trained using Almost No Inner-Loop \ac{MAML} (ANI-MAML)  \cite{Raghu2020Rapid} to adapt to unseen tasks easily. 
The AFR module uses an attention mechanism to identify the most relevant past features and selectively reuse them in the current task. The attention weights are learned jointly with the other model parameters during training, allowing the model to adapt to the task and identify the relevant features on the fly.
Therefore, the attention mechanism and meta-learning enabled task heads to learn to adapt to new unseen tasks in fewer data instances.
This work evaluates the proposed method for the tasks of image classification and estimation of depth, vanishing point, and surface normal from a single input image.
The key difficulty was to learn task-invariant representations in addition to task-specific representations.
This was addressed by the \ac{MTL} design, which provided inductive bias on selected features as specified by the attention technique.

 \cite{upadhyay2023multitask} proposed an alternative approach to integrating multi-task learning (MTL) and meta-learning. Their method involves the simultaneous learning of multiple diverse, dense pixel-level tasks, with the aim of facilitating the seamless adaptation of a new task in a multi-task context.
The approach implemented in this study involves the utilization of multi-task learning episodes, which consist of combinations of multiple tasks that are trained through a bi-level optimization scheme based on the \ac{MAML} framework.
The authors showcase their methodology by applying it to two distinct datasets, namely NYU-v2  \cite{Silberman:ECCV12} and taskonomy  \cite{zamir2018taskonomy}. The four tasks evaluated in this study include semantic segmentation, depth estimation, surface normal estimation, and edge detection.
The present study conducted a comprehensive comparative examination of single-task learning, (vanilla) multi-task learning, and multi-task meta-learning. 
Additionally, the study evaluated the performance of these approaches when a new task is introduced. 
The findings indicate that the integration of a new and unfamiliar task yields superior and more efficient results in multi-task meta-learning as compared to multi-task learning.
The literature presented above encompasses a variety of works that integrate \ac{MTL} and meta-learning.
Categorizing these works based on their approach or novelty proves to be a challenging task, given their substantial heterogeneity.

The article  \cite{Yang_2022} discussed the use of graph neural networks and multi-task meta-learning for analyzing brain connectomes.
In this work, the multiple tasks refer to the different modalities from the dataset.
The authors propose a data-efficient learning framework that leverages meta-learning techniques (MAML, in particular) and brain network-oriented design considerations to make predictions about brain network-based diseases. The framework extends single-task training into multi-task training by expanding the pre-training phase into simultaneously co-learning over multiple source objectives. This approach allows for more efficient use of available data and can help to overcome issues related to overfitting and generalization.


\subsection*{Some Exceptions:} Here are discussed some works that claim to combine meta-learning and \ac{MTL} together but do not fall under the aforementioned categories.
The title of the article by  \cite{wang2021bridging} gives an impression of integrating meta-learning and \ac{MTL} by mentioning the term `bridging' the two learning paradigms.
Although they have demonstrated both conceptually and empirically that \ac{MTL} is a computationally viable approach to gradient-based meta-learning algorithms, this is especially true for sufficiently deep networks.
This is because the learned predictive functions of \ac{MTL} and meta-learning are similar and share similar optimization objectives.
Another work by  \cite{abdollahzadeh2021revisit} presented multi-modal meta-learning for only one classification task while using the transference metric  \cite{fifty2021measuring} of \ac{MTL} to update shared parameters. 
In  \cite{leaving}, a multi-task meta-learning approach is employed for learning multiple scenarios in a single model for an advertising problem. 
They proposed a meta unit that transforms scenario information into dynamic weights \ie a network of fully connected layers; this meta unit is connected to the network architecture claiming to in-cooperate meta-learning along with multi-task (multi-scenario) learning. 
There are also a few articles such as  \cite{pmlr-v119-bronskill20a, ghadirzadeh2021bayesian, kedia-chinthakindi-2021-keep, Tian2019HierarchicalIN, 8769969, krueger2020hidden, 10557138, gu2024proto} which by abuse of terminology refer to the training of multiple tasks or learning episodes in meta training stage as \ac{MTL}, thereby considering the work as meta multi-task learning (or multi-task meta-learning).
According to the definitions of the learning paradigms in Sec.~\ref{learning_paradigms} of this article, training multiple tasks jointly is different from training multiple learning episodes sequentially. Therefore, such works are not added as related literature in this section.
A handful of articles by  \cite{Retyk_2021, cs6587, ghosh2020deployment, JMLR:v10:li09b} that focus on meta multi-task reinforcement learning are not discussed here as they are beyond the scope of this work.

\subsection{Research in multi-task transfer learning} \label{MTL_transfer}
Transfer learning and \ac{MTL} usually differ in how and when the information is shared between tasks. 
Both can handle heterogeneous tasks, and both share model parameters and features between tasks.
Since the underlying theory behind \ac{MTL} is an inductive transfer approach, it is considered to be a derived form of transfer learning  \cite{Pan10asurvey}, but they are very different, as discussed in Sec.~\ref{comparisons}.
Various combinations of these learning paradigms are possible, and a few examples are provided below.
In the context of multi-task transfer learning, it is common practice to train a multi-task model to address multiple tasks within a given domain. Subsequently, this trained model is leveraged to address the same tasks within a distinct domain. 
The model obtained from a particular source is subjected to fine-tuning using data from the target domain, with the aim of acquiring the ability to perform multiple tasks together. 
While in some of these works the source and target tasks are different, the source and target domains share a portion of the multi-task model.

\begin{figure}[ht]
    \centering
    \includegraphics[width = \linewidth]{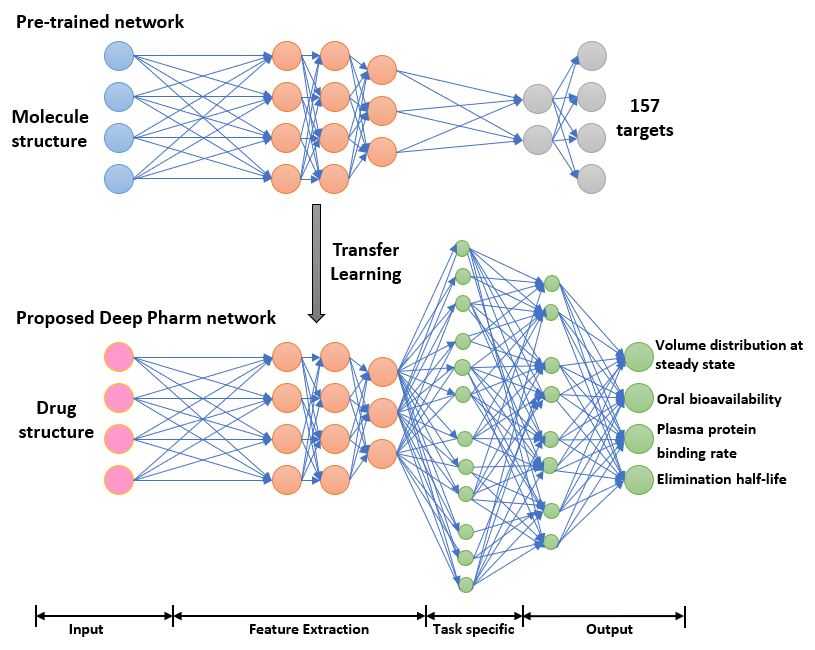}
    \caption{Illustration of the integration of transfer learning and \ac{MTL}  \cite{Ye_2018}.}
    \label{fig:deepPharm}
\end{figure}

 \cite{Ye_2018} aimed to predict pharmacokinetic parameters by learning a model for quantitative structural activity relationships of drugs.
The four parameters, i.e., oral bioavailability, plasma protein binding rate, apparent volume of distribution in the steady state, and elimination half-life, were estimated in this work. 
These four parameters are considered multiple tasks. 
First, a pretrained model is learned on an extensive bioactivity dataset, the knowledge from which is further used in the multi-task DeepPharm model proposed in the article, as depicted in Fig. \ref{fig:deepPharm}.
The integrated multi-task and transfer learning approach enhanced the generalization of the model compared to the conventional models and overcame the lack of sufficient and high-quality data in the \ac{ADME} evaluation. 
Such a well-generalized model may be helpful to perform \ac{ADME} calculations on the new drug structures using the DeepPharm model, \ie making inferences only, as there is no need to train the model again on different data.
This article opens a new dimension of research drug discovery and development using combining \ac{DL} algorithms.

An example of another type of fusion, \ie when the source task is \ac{MTL} from which the learned information is transferred to a target task, is illustrated in an article by  \cite{s20247205}.
The aim of this article is fault diagnosis of rolling element bearings under uncertain working conditions.
First, \ac{MTL} is enforced to determine the speed and health type of the machines (in general) by using the bi-spectrum-based analysis of the vibration signals as inputs. 
Furthermore, transfer learning is used to enhance the classification performance by using the proposed \ac{MTL}-\ac{CNN} as the source task and identifying the bearing faults under severe conditions as the target task.  
Since there are usually fewer faulty bearing data in extreme conditions (as this is a type of anomaly), the transfer learning technique is advantageous.
The pre-trained model for extracting features in the target dataset in a multi-task setting leads to good performance on the unseen target data. 

In  \cite{cruz-etal-2020-localization}, pretrained transfer learning models such as BERT  \cite{devlin2019bert}, ULMFiT  \cite{howard2018universal} and  GPT-2  \cite{radford2019language}, were used for fake news detection. 
This included an auxiliary language modeling task to adapt to the writing style of the downstream task of fake news detection.
Therefore, multi-task fine-tuning was introduced in this work by combining losses from both tasks. 
In a similar work by  \cite{dong_0216046}, transfer learning and \ac{MTL} were used together for \ac{NER} on Chinese \ac{EMR}.
This trains a bi-directional \ac{LSTM} in general (source) domain and further uses the acquired knowledge from the source domain to improve the performance of \ac{NER} in Chinese EMR, \ie the target domain.
\ac{POS} tagging and \ac{NER} are the two tasks in the target domain that are trained alternatively so that knowledge from one task may enhance the knowledge gained by the other task.
Since this work aims at \ac{NER}, \ac{POS} is treated as an auxiliary task that aids in better learning of Chinese \ac{NER}. 
On similar grounds,  \cite{taslimipoor_cross} intends to classify the multi-word expressions with the help of two auxiliary tasks of dependency arcs and labels.
To attain this, it utilizes knowledge shared by the pretrained model for \ac{POS} and dependency parse tags for two different languages.
This is because the target language has limited resources, and cross-lingual transfer learning helps overcome this issue.

Multi-task transfer learning is also used in  \cite{9229132} for neural decoding, which decodes the brain activity to reconstruct visual information. 
To achieve this, first, fMRI voxel features of the brain are decoded into \ac{CNN} features by using \ac{SMR} ( \ie Voxel2Unit).
Then, the predicted \ac{CNN} features are converted to an image using introspective conditional generation (\ie Unit2Pixel).
Multiple \ac{CNN} features are decoded from the FMRI data using SMR, and every single output prediction is considered a task, thereby applying \ac{MTL} for the Voxel2Unit process.
For the Unit2Pixel process, a pretrained \ac{CNN} called AlexNet  \cite{alexnet}, trained on large image data ImageNet  \cite{5206848} is used for image reconstruction as a part of deep generative models, particularly a combination of \ac{VAE}  \cite{kingma2014autoencoding} and \ac{GAN}  \cite{goodfellow2014generative}.

The article  \cite{Nguyen_2017} analyzed the characters of a digital Japanese comic named Manga, using Bi-modal inputs, \ie the graphics and text information. 
This approach uses pretrained networks such as BERT  \cite{devlin2019bert}, and ResNet  \cite{he2015deep} for text and visual feature extractors, respectively, to obtain the feature embeddings to combine data from both modes. 
These embeddings are further used in a multi-task architecture for character retrieval, identification, and clustering tasks.
These three tasks are independent; each can be accomplished by training three different models using the same images and text (uni-modal inputs can also be used).
However, since they share common multi-modal inputs (image and text), they are suitable for \ac{MTL}. 
Certainly, sharing parameters between tasks improves the performance of all tasks compared to when they are trained in isolation.  
The strength of \ac{MTL} \ie handling multi-modal inputs and multiple tasks and transfer learning \ie good feature representations, is demonstrated in this work. 

 \cite{8920992}, exploited a  few-shot Dirichlet-Net based \ac{MTL} for hyperspectral image classification. 
An encoder-decoder network to reconstruct hyperspectral images was used, and the representations were shared using the encoder in a classifier network. 
This ensemble of encoder-decoder and classifier is termed as \ac{MTL} in this article.
The encoder takes two images of different domains as input and extracts both representative and discriminative vector representations from both domains; these encoder network parameters are shared with the classification network to extract features of the input image patches to predict output labels.
The important contribution of this article is the extraction of shared representations from objects in different domains by applying transfer learning.

Another exciting application by  \cite{Dong2021} aimed to estimate virus-human protein interactions by pre-training a source model called UniRep  \cite{Alley589333} to produce features or protein embeddings.
Due to the scarcity of training data for virus-human protein interactions, the article trained a network on powerful statistical protein representations, \ie source tasks.
The target task was performed \ac{MTL} by extracting the embeddings using the source network to find human protein-protein interactions (PPIs) and human virus PPI. 
Likewise,  \cite{8404496} used VGGNet  \cite{simonyan2015deep} for pre-training, and the extracted features were used to detect and identify copper and plastic wires buried in the ground using  Ground Penetrating radar (GPR) scans.
Identifying the type of soil (wet or dry) in the target domain is also added, making it a \ac{MTL} application.
It is observed that in many of the articles, such as   \cite{cruz-etal-2020-localization, Dong2021, simonyan2015deep}, a source network is trained on a large dataset of one domain, which is exploited by the target dataset of another domain having fewer data resources.
Therefore, transfer learning helps to find good feature representations, leading to better generalization.  
Along with the multi-task architecture, it makes it possible to train many tasks jointly.

A similar work  \cite{xiao2024integrated} developed an integrated transfer learning and multitask learning strategy using graph neural networks (GNNs) to predict bioaccumulation parameters of chemicals. 
Transfer learning pre-trains the model on n-octanol/water partition coefficient data, and multitask learning enhances performance by training on multiple bioaccumulation parameters simultaneously. 
The TL-MTL-GNN model significantly outperformed single-task models and conventional machine learning methods, demonstrating robust accuracy and efficiency, especially with limited data. 
This approach effectively addresses small data challenges in environmental science, predicting bioaccumulation for approximately 60,000 chemicals.

\textit{Some Exceptions}: 
 \cite{Xu2011ASO}, give a detailed survey of how transfer learning and \ac{MTL} individually are used in bio-informatics.
 \cite{Kamath2019} is a book chapter that discusses transfer learning types and describes MTL as a variant of transfer learning. 
While  \cite{Sun2019, RICCI201767} details types of multi-view transfer learning and multi-view \ac{MTL}.
The articles  \cite{Wang_Pineau_2015, maurer2014sparse} give theoretical concepts of improving the performance of \ac{MTL} and transfer learning in general.

\subsection{Research in Meta transfer learning}
Transfer learning and meta-learning are extremely similar in the sense that they both involve source tasks and unknown target tasks, and the goal of each is to achieve a superior generalization on the target task.
Meta-learning, in contrast to transfer learning, has a two-level optimization technique.
Transfer learning is now ubiquitous, as pretrained models are used in nearly all applications and algorithm types; they are either fine-tuned or trained from scratch for various applications.
However, numerous studies in the literature claim to use a combination of transfer learning and meta-learning, and a few of these are mentioned here.

As already discussed in Sec. \ref{MTL_transfer}, transfer learning can be crucial in extracting data features using a pre-trained model and, therefore, can integrate with other learning algorithms.
The article by  \cite{Sun_2019_CVPR, 9173698} followed a similar approach for performing few-shot learning. 
Transfer learning is used during meta training by extracting representations of images using a model trained on a very large dataset such as MiniImageNet  \cite{NIPS2016_90e13578}.
The meta transfer learning proposed in this work answers two critical questions, \ie what to transfer and how to transfer.
The \ac{DNN} parameters trained on the large-scale data answer what to transfer, whereas the scaling and shifting operations learned for each task introduced in this work refer to how to transfer.
The meta-knowledge from the training stage is shared in the testing stage on the Fewshot-CIFAR 100  \cite{FC100} dataset to learn new tasks using fewer data instances.
They also propose a hard task meta-batch strategy in which, rather than randomly picking meta-training tasks, the algorithm resamples the hard tasks based on past validation accuracy and failure.  
The use of pre-trained \ac{DNN} was proved to be very useful for tailoring the learning experience for unseen tasks.

Likewise,  \cite{Soh_2020_CVPR} use transfer learning together with the optimization-based meta-learning method \ac{MAML}  \cite{pmlr-v70-finn17a} for zero-shot super-resolution of images.
For faster adaptation and better generalization on new tasks, meta-learning helps learn effective initial parameters during training.
At the time of meta-testing, it takes only a few gradient steps to learn the image-specific information, even in the case of external (or new) data instances.
The learning strategy introduced in this work learns initialization parameters with reference to different blur conditions, making it possible to adapt to new (unseen) blur kernels quickly.
meta-learning is used to learn image-specific (internal) information, while transfer learning is adopted prior to meta-learning to utilize the external samples.
The learning episodes during meta-training were images from different blur kernels that helped to adapt the image-specific features quickly. 
According to the authors, the proposed method can be interpreted as a self-supervised approach to super-resolution.

A recent article by  \cite{Willard_2021} adopted meta-transfer learning for predicting the dynamics of the water temperature of unmonitored lakes.
Since the data for the unmonitored lakes are insufficient and all deep learning models need a significant amount of data to learn, this work utilized the available data of the monitored lakes to learn the models and transfer knowledge to the domain with fewer resources. 
It exploits the monitored lake data by extracting important characteristics and using two source models, i.e., process-based and process-guided deep learning, on each monitored lake and evaluating the model performance, thereby applying transfer learning to the unmonitored lakes.
This meta-knowledge of features and performance of the models of the monitored lakes is used to select the best model for the unmonitored lakes based on the lowest predicted error.
These pre-trained models outperform on the target unmonitored lake compared to when no transfer of learning is performed.
A very similar work by  \cite{water_temp} also proposed a meta-transfer learning method for predicting daily maximum water temperatures in stream network, emphasizing the accurate forecasting of extreme temperature events. 
Their approach uses transfer learning to fine-tune models with historical data and meta-learning to reweight samples based on similarity
They also incorporated an extreme value loss function and clustering to improve accuracy and efficiency. 
This method significantly enhances predictive performance for both normal and extreme periods.

For vehicle tracking using Unmanned Aerial Vehicle (UAV)  \cite{Song_UAV}, used a pretrained model for vehicle tracking on ground images and employed the model to adapt to the drone view images.
From a deep learning viewpoint, vehicle tracking in UAVs is an under-explored research area, as these videos have significantly less labeled data.
Transfer learning is employed to overcome this problem of data availability.
A large ground view vehicle tracking dataset is used to train a model, which is then used by the drone-view dataset, therefore transferring features across landscapes.
meta-learning is used to adaptively extract shared features between both domains (drone and ground view).  
Therefore, transfer learning helps to overcome the data scarcity problem, while meta-learning solves the issue of domain shift. 

 \cite{10506110} employs transfer learning by using a pretrained model on large-scale natural Hyper Spectral Image (HSI) datasets and adapts it to remote sensing HSI datasets via meta-learning, which involves learning task-specific adjustments with scaling and shifting operations for hyperspectral image super-resolution. 
The meta-training process involves creating multiple tasks by randomly sampling continuous spectral bands from the hyperspectral images. Each task represents a super-resolution (SR) problem with different spectral ranges. 
By training on these diverse tasks, the model learns general super-resolution patterns that are not specific to any particular spectral range but are instead broadly applicable to a variety of spectral configurations. 
This results in a robust and adaptable framework for hyperspectral image super-resolution, which solves two key issues, i.e., data scarcity and domain differences in the case of HSI datasets.

A meta-transfer learning method for gearbox fault diagnosis with limited data is introduced in  \cite{She_2024}.
This method used transfer learning to align data distributions between different operating conditions (source and target domains), such as varying speeds and loads, to simulate real-world scenarios., enabling the model to leverage knowledge from one condition to diagnose faults in another. 
Meta-learning was employed to dynamically adjust the model parameters using meta-stochastic gradient descent (Meta-SGD) \cite{li2017meta}, which allowed for faster adaptation and improved generalization across tasks with less data. 
The method demonstrates high accuracy in fault diagnosis with limited samples by constructing subtasks and aligning domain distributions using adaptive manifold regularization.
Each subtask included data for various fault types of the gearbox, such as tooth breakage, tooth pitting, and compound faults.

 \cite{wu-etal-2023-good} in their study introduced a Meta-Task Collector-based Cross-lingual Meta-Transfer framework (MeTaCo-XMT) for enhancing few-shot learning in low-resource languages by leveraging high-resource language data through transfer learning i.e., fine-tuning pre-trained multilingual models such as mBERT  \cite{devlin2019bert} and XLM-R  \cite{conneau2020unsupervised} on high-resource language data (e.g., English). 
They employed meta-learning to construct and optimize meta-tasks using various data selection strategies, including syntactic similarity sampling.
This approach significantly improved cross-lingual transfer performance on NER and MRC tasks, particularly for low-resource languages such as Swahili, Afrikaans, Tagalog, Javanese, and Yoruba. 
Their contributions include the development of a syntactic distance metric model and demonstrating robust, high-accuracy performance with minimal data.

\textit{Some Exceptions}:  The `transfer of meta information' (which is solely meta-learning) is also termed as meta-transfer learning in articles such as  \cite{8545411, bastani2021meta, duan2021ultra, pmlr-v27-aiolli12a, 9839571, 10.1007/978-981-19-4549-6_18}.
Like  \cite{8545411}, the meta-information for facial emotion classification on one dataset is transferred to another dataset.
Similarly,  \cite{9839571} discusses cross-subject EEG emotion recognition tasks. It uses meta-learning to adapt to out-of-distribution data \ie for subjects with less (or insufficient) data.
 \cite{bastani2021meta} discuss transferring knowledge across experiments for dynamic pricing applications. 
 \cite{pmlr-v27-aiolli12a} introduced a variant of meta-learning, which learns how to learn kernels from data and share the sequence of transformations to find a kernel for a new task.  
These are solely meta-learning applications, but due to linguistic inconsistencies, they fall under the category of meta-transfer learning when actually it refers to simply the transfer of meta-learnings (or knowledge).
This combination of transfer learning and meta-learning enabled the model to generalize better and make more accurate predictions for low-resource languages, even with limited annotated data.


\section{Discussion} \label{discussion}
\begin{figure*}[t]
     \centering
     \includegraphics[width = 0.7\textwidth]{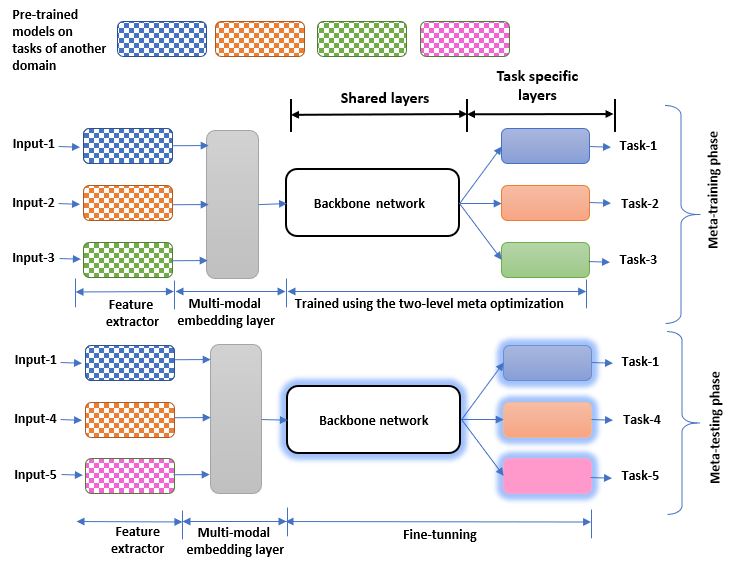}
     \caption{Proposed implementation, that ensembles \ac{MTL}, meta-learning and transfer learning. Here the checkered rectangle boxes represent the pretrained layers (or model), the solid rectangle boxes are the layers trained from scratch, and the glowing rectangle boxes denote the layers which are fine-tuned, \ie the parameters from the source model act as initialization for the target network.}
     \label{fig:P1}
 \end{figure*}

Inspired by the existing literature on the fusion of two algorithms as discussed in Sec.~\ref{ensemble}, we introduce \textit{\ac{MMMTL}} as the combination of all the three learning paradigms, \ie \ac{MTL}, meta-learning and transfer learning.
Multiple approaches may exist for the implementation of this ensemble. 
The illustration presented in  Fig.~\ref{fig:P1} is an instance of a highly generic execution of the aforementioned.
Based on the features of these algorithms, the primary objectives of this ensemble are as follows:
\begin{enumerate}
    \item Good performance on a new unseen task or data (due to meta-learning)
    \item Ability to handle multi-modal inputs and heterogeneous tasks (due to \ac{MTL})
    \item Require less training data and good feature representation for learning (due to transfer learning)
\end{enumerate}
 
The approach proposed in Fig.~\ref{fig:P1} entails a fusion of representational knowledge transfer and functional knowledge transfer. 
The architecture of the model is in accordance with \ac{MTL}; the backbone network and multi-head modules are trained using a two-level optimization that is employed in meta-learning.
Assuming the inputs are multimodal, the pre-trained networks (checkered boxes) extract essential features from the inputs, and these are processed accordingly in the multimodal embedding layer before forwarding to the backbone network, similar to what was presented by  \cite{zhang2021multi,openended}.
The backbone network refers to the generic layers of the neural network or \ac{CNN}, whose input combines embeddings from various inputs.
The output from the backbone network goes into the task-specific layers.
The architecture of every task-specific layer can differ as required by the task. 
Therefore, in the meta training stage, the multi-task architecture helps to learn many tasks together, thereby improving the performance of all the tasks.
When new tasks (Task-4 and Task-5) are introduced in the meta-testing stage, it is easy to learn them faster and in fewer data instances. 
The inductive bias of the multi-task architecture assists in better generalization than single-task learning.
Also, this stage allows for both feature extraction and fine-tuning variants of transfer learning.
This proposed implementation gives the liberty to add new heterogeneous tasks and allows for multi-modal inputs; transfer learning and meta-learning also enable learning with fewer data samples during meta testing.
Based on the theoretical properties of the three learning paradigms (discussed in Section II), we hypothesize that \ac{MMMTL} could potentially improve the performance of both the source and target tasks, thereby aligning with the three objectives mentioned above.
 \begin{figure}[ht]
     \centering
     \includegraphics[width = \linewidth]{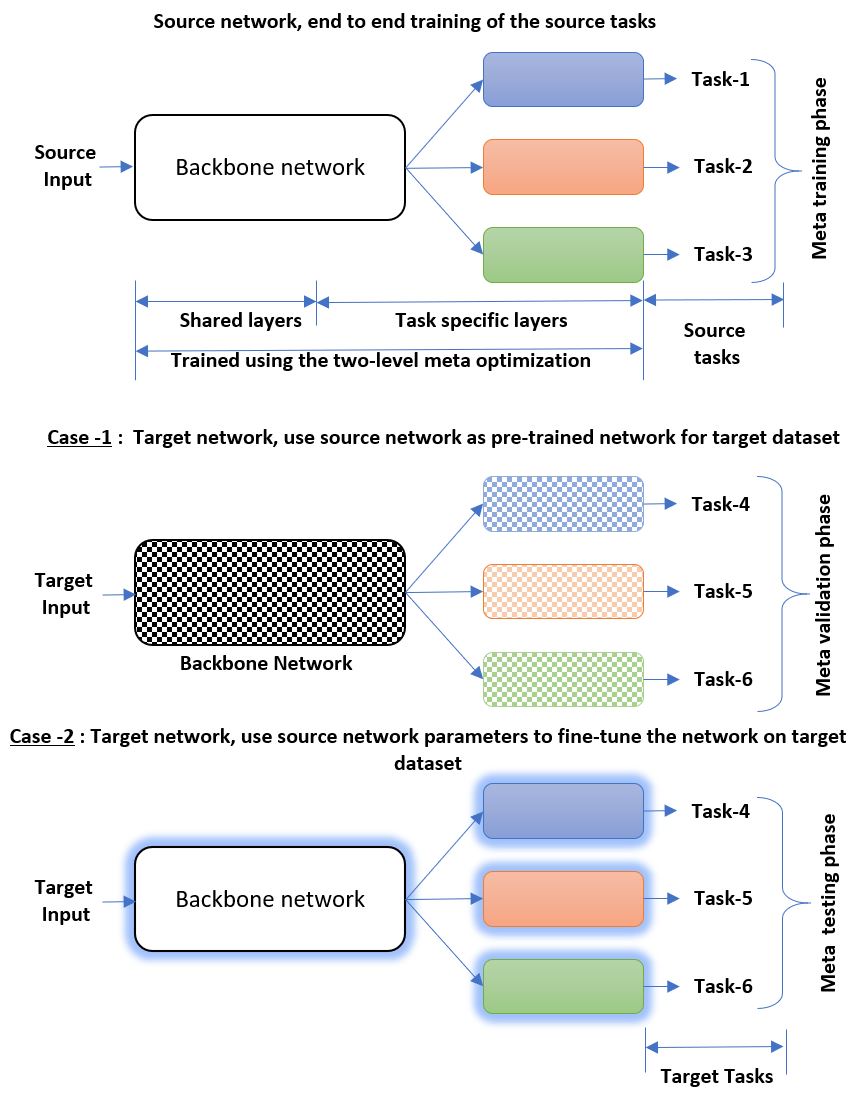}
     \caption{Variant of the proposed implementation in Fig.~\ref{fig:P1}, showing instance of single input and multiple tasks. Case-1 and 2 can be understood from two points of view: transfer learning and meta-learning, while architecture represents \ac{MTL}. Here, checkered rectangle boxes represent pretrained layers (or models), solid rectangular boxes are layers trained from scratch, and glowing rectangle boxes denote layers that are fine-tuned for unseen target data.}
     \label{fig:P2}
 \end{figure}
 
A variant of the implementation in Fig.~\ref{fig:P1} is shown in Fig.~\ref{fig:P2}.
It represents the instance when there is one input and many tasks related to it (similar to the illustration in Fig.~\ref{fig:PadNet}).
Here, a multi-task source architecture can be meta-trained for many different datasets and similar tasks.
Later, the network is used for some data of a different domain, and the meta information from the training phase is helpful for improved learning of the target tasks.
The trained network on the source data can be employed as a feature extractor \ie Case -1 in Fig.~\ref{fig:P2} or fine-tuned for the target data \ie Case-2 in Fig.~\ref{fig:P2}, is the transfer learning point of view.
Another way of viewing this approach is from the meta-learning point of view, wherein the training of the source tasks can be considered as the meta training stage, and case-1 can be treated as the meta-validation stage because there is no training.
It is only making inferences on new data, and case 2 is nothing but the meta-testing stage, which uses the initial parameters from the source model and then trains the network on a new target dataset.

Fig.~\ref{fig:P1} is a plausible approach to how to fuse the learning paradigms.
While this approach adheres to the three primary objectives, it is important to acknowledge that there are likely several other viable methods to achieve this integration. 
Each approach should aim to leverage the advantages of these learning algorithms while addressing their respective weaknesses. 
Also, these approaches depend on the use case and the available dataset, \ie the number of tasks, modality of the data, amount of data samples, etc. 
To the best of our knowledge, there is no such work in the literature that employs the ensemble of meta-learning, transfer learning, and \ac{MTL}.
Indeed, the related work or future scope section of many articles mentions these three learning algorithms, but none illustrate applying them together.

\subsection{Practical example}\label{example}
For better understanding, here we very broadly present a real-world example where the proposed \ac{MMMTL} can be applied.
Consider a scenario of autonomous vehicle navigation through diverse environments that need to understand its surroundings using multiple sensor inputs. 
These input modalities include RGB images and LiDAR point clouds. 
The aim is to perform dense prediction tasks, which include;
\begin{enumerate}
    \item Semantic segmentation: Classifying each pixel of an image into predefined categories, for e.g., road, pedestrian, and vehicle. 
    \item Depth estimation: Estimating the distance of each pixel from the sensor. 
    \item Surface Normal estimation: Determining the orientation of the surfaces in the scenes. 
\end{enumerate}

\subsubsection*{Implementation of \ac{MMMTL}}
\textbf{Step 1: Transfer Learning} This includes feature extraction using pre-trained models of the respective modality.
The pre-trained models can be trained on tasks that are similar to the ones mentioned earlier, or they can be trained on different source tasks using RGB or LiDar input data from a similar (autonomous driving) or different domain.  
These extracted embeddings capture essential features and representations from the source domain. 
However, we can also fine-tune these models further to better adapt the embeddings for our tasks.  
In the Fig.~\ref{fig:P1}, this step is represented by the checkered boxes. \\

\textbf{Step 2: Multi-modal embedding layer} This is represented by the grey box in Fig.~\ref{fig:P1}, which is responsible for exploiting the embeddings from different modalities to accomplish the tasks at hand. 
There are multiple approaches to how many modalities can be handled or fuse together or condition each other. 
 \cite{huang2022multi, TANG2023109165, 9000872} provide valuable insights into this area of research.\\

\textbf{Step 3: Multi-Task learning} The resulting embeddings from Step 2 can be used to train a multi-task network, that consists of a shared network and the respective task-specific heads (or decoders). 
Fig.~\ref{fig:P1} shows a hard parameter-sharing approach using the white block as the backbone network and the colored solid blocks as the decoders for semantic segmentation, depth estimation, and surface normal estimation. 
 For instance, a ResNet50  \cite{he2015deep} can be used as the shared network, and DeeplabV3  \cite{chen2017rethinking} for the task-specific heads. 
This is merely one of the approaches to achieving multi-task learning, there are several other architectures in the literature that can also be employed in this step.\\

\textbf{Step 4: Meta learning} During meta-training, the learning episodes can be created by sampling tasks and data from different environments (urban, suburban, and rural scenes).
Use bi-level meta-optimization to adapt the shared backbone and task-specific heads, ensuring quick adaptation to new tasks and environments, similar to MAML \cite{pmlr-v70-finn17a}.\\

This is how these three learning paradigms can be assembled to achieve \ac{MMMTL}.
Additionally, there are other factors to be considered in the ensemble that are not addressed in this paper.
These factors include the selection of loss functions for different tasks, task balancing in multi-task learning, and hyper-parameter tuning for stable bi-level optimization, among others.
Theoretically, based on the properties of these individual paradigms, the combination should benefit from (i) \textit{improved generalization} due to the shared backbone and fused features from multiple modalities across different environments, (ii)\textit{ data efficiency} because of meta-learning that allows the network to adapt to new tasks with limited data, and (iii) \textit{integration of many modalities} help to leverage complementary information, leading to inter-modality and inter-task knowledge transfer.

\subsection{Open questions}\label{questions}
The proposed approach, known as \ie \ac{MMMTL}, is a generic ensemble of three distinct learning algorithms: Multi-Task Learning (MTL), meta-learning, and transfer learning. Theoretically, this approach has the potential to address many of the limitations associated with using these algorithms in isolation.
As demonstrated previously, there may exist multiple variations of \ac{MMMTL}.
This survey article enlists a few open research questions pertaining to these ensembles, with the aim of facilitating a comprehensive empirical exploration of this approach.
These questions are presented under distinct headings corresponding to the diverse attributes of the ensemble.\\
\noindent
\textbf{Feasibility}
    \begin{itemize}
    \item[1] In what ways, \ac{MTL}, meta-learning and transfer learning be integrated into a single \ac{MMMTL} framework?       
    \end{itemize}
\textbf{Performance}   
\begin{itemize}
    \item[2] How is the proposed \ac{MMMTL} approach better than single task learning?
\end{itemize}
\textbf{Modularity and self-sufficiency}
\begin{itemize}
    \item[3]  In \ac{MMMTL}, is there a possibility for modular learning? In particular, for an unseen task, can the network automatically choose, based on the meta-knowledge, which part of the network should be trained rather than training the whole network and how to exploit the shared structure (using transfer learning)?
    \item[4] Do similar tasks automatically resort to similar sub-architecture training?
    \item[5] How can meta-learning and transfer learning help to automatically alter the network architecture for a new task?
\end{itemize}
\textbf{Multi-modality}
\begin{itemize}
    \item[6] In the case of multi-modal inputs, analyze the contribution of each modality to the outcome.
\end{itemize}

\subsection{Some limitations}

The present article centers on an integration of knowledge-sharing algorithms with the aim of mitigating the limitations of one algorithm through the strengths of another.
However, there exist certain constraints that are insurmountable and persist as an obstacle to be resolved.
For example, As the number of tasks in the multi-task environment grows, the size of the model increases; as a result, the number of parameters to be trained also increases. Therefore, \ac{MTL} is computationally expensive.
Similarly, with meta-learning, the two-level optimization is effective but is both time and memory expensive. 
Unfortunately, although highly beneficial, the computational cost of such complex fusion models like \ac{MMMTL} is significant.

In \ac{MTL}, diverse tasks may cause negative transfer.
The definition of diverse or distinct tasks is subjective, as no metric measures the similarity or dissimilarity between the tasks.
Moreover, it depends on the use case and the tasks at hand to tag them as similar or diverse tasks. 
If the tasks are very different, there is a threat of negative information transfer between the tasks.
One way to solve the problem of negative information transfer is by employing better task-balancing approaches and task-specific hyper-parameters.
If there is a significant disparity in the loss values across tasks, task balancing can help by equalizing the gradients associated with each task. 
Soft parameter sharing can also be a way to overcome this issue.
Negative transfer between the source and the target domain is also a problem in transfer learning.

The gradient-based meta-learning algorithms follow an episodic learning approach during training, wherein small episodes or, in simple words, distinct tasks are used to train a model.
A major issue arises while combining meta-learning with multi-task learning: How do we create learning episodes?
The article by  \cite{upadhyay2023multitask} suggests a way to create multi-task learning episodes to meta-train a multi-task model.
However, if there are more than 5-6 tasks, the suggested method is very computationally expensive.
In conclusion, identifying how to create learning episodes to implement meta-learning in a multi-task and multi-model setting poses a significant challenge.

\section{Conclusion and future work}
This article provides an overview of transfer learning, \ac{MTL}, and meta-learning. 
It presents a comparative analysis of these three algorithms and reviews the existing literature on their combined use. 
Additionally, the article proposes an approach \ac{MMMTL} for integrating all three learning paradigms to mitigate some of their respective limitations.
The suggested learning network may be regarded as a generic learning framework, as it can be deconstructed into its constituent algorithms and various combinations thereof.

Since the global network introduced in this work makes it possible to choose the elements required to learn the task, it will undoubtedly be worthwhile to meta-train the network to learn to evaluate which elements will do justice to a task rather than employing all the learning algorithms. 
Because there is a possibility that jointly learning some tasks is not possible (because of negative transfer), and in such a scenario, the idea is to learn to activate only parts (modules) of the network, thereby foreseeing a formulation of the meta-meta-learning algorithm.
In consequence, as future research, it will be interesting to explore how these learning paradigms together share to learn and learn to share, along with knowing when to share, thereby making it possible to develop more human-like learning techniques.
In summary, combining these knowledge-sharing algorithms has the potential to catalyze noteworthy advancements in the field of deep learning.

\appendices
\section*{Acknowledgment}
The authors express profound gratitude to Prof. Atsuto Maki from KTH Royal Institute of Technology for his invaluable comments on the licentiate thesis\footnote{In Swedish and Finnish universities, a licentiate, which is a pre-doctoral degree, is achieved after two years of full-time research studies and equals 120 credits. It involves completing coursework and a dissertation, which is formally half of a doctoral dissertation. The licentiate can be a step toward a PhD or a final study goal, and the main work called a licentiate thesis, is presented and evaluated at a public seminar.} from which this work has evolved. 
Additionally, we extend our heartfelt thanks to all reviewers who have meticulously reviewed previous versions of this article and provided insightful inputs. We have endeavored to integrate all constructive comments to enrich this version of the article.

\bibliographystyle{IEEEtran}
\bibliography{references}

\begin{IEEEbiography}[{\includegraphics[width=1in,height=1.25in,clip,keepaspectratio]{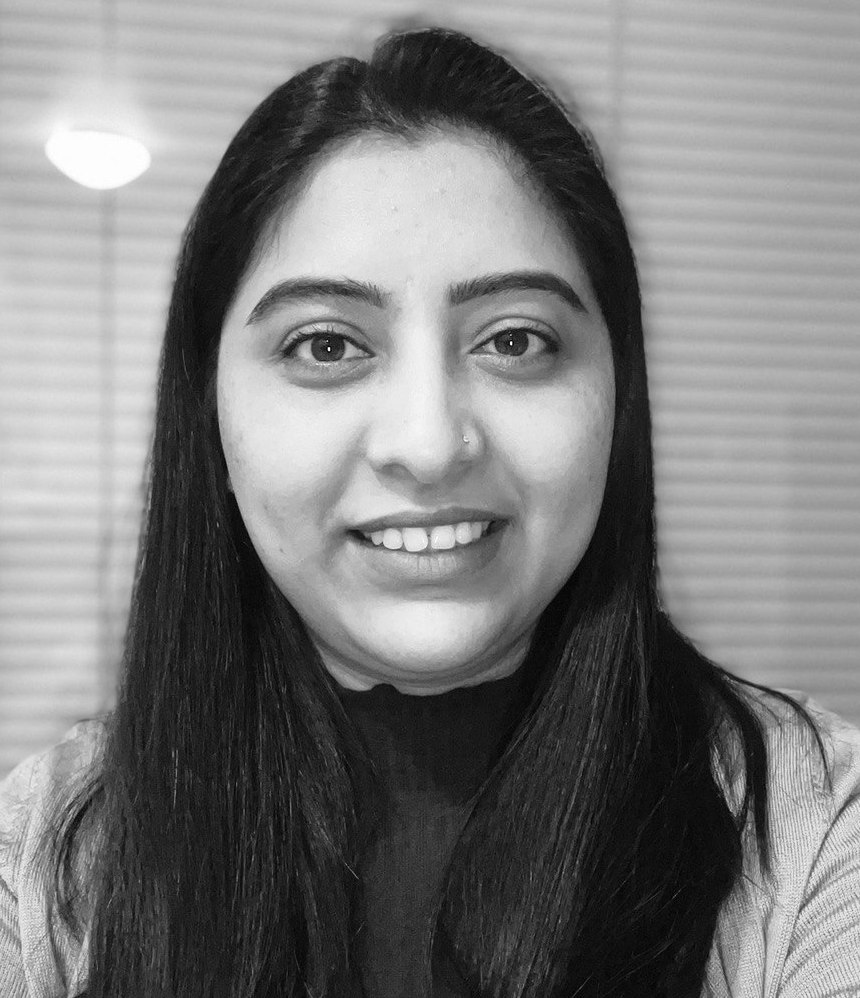}}]{Richa Upadhyay}{\space} is PhD researcher currently associated with the Machine Learning group at the Luleå University of Technology, since October 2020.
She earned her Master of Science (MS) degree in Signal and Image processing Methods and Applications (SIGMA) from the Institut polytechnique de Grenoble (Grenoble INP), France, in 2019. Prior to this, she obtained her Master of Technology (M.Tech) in Electronics from the Veermata Jijabai Technological Institute, Mumbai, India, in 2013.
Her current research interests encompass several areas, including Multi-Task Learning, meta-learning, Neuroscience, Computer Vision, and Signal \& Image Processing.

    \end{IEEEbiography}
    
    \begin{IEEEbiography}[{\includegraphics[width=1in,height=1.25in,clip,keepaspectratio]{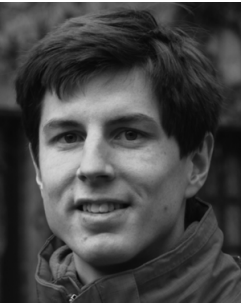}}]{Ronald Phlypo}{\space} received the M.Eng. degree in electrical engineering from KHBO, Ostend, Belgium, in 2003,
    the M.Sc. degree in artificial intelligence from
    KU Leuven, Leuven, Belgium, in 2004, and the
    Ph.D. degree in engineering from the University of
    Ghent, Belgium, in 2009. From 2009 to 2012, he was
    with the GIPSA lab, Grenoble, France, as a PostDoctoral Research Associate, where he was involved
    in a signal processing model for joint Gaze-EEG
    recordings. From 2012 to 2013, he was an Associate
    Researcher with the Machine Learning for Signal Processing Lab, Baltimore, MD, USA, where he was involved in linear latent variable models
    for multimodality. In 2013, he has held an engineering position with the
    INRIA Institute, Saclay, France, where he was involved in the untanglement
    of functional magnetic resonance imaging cognitive connectivity estimates.
    Since 2014, he has been an associate professor position with PHELMA,
    Grenoble Institute of technology, Grenoble, France, with a research component with Saint-Martin-d’Hères, Grenoble. His current research interests are
    on functional connectivity estimates, multimodality, and multistability with
    applications to biomedical signal processing. 
    \end{IEEEbiography}
    
    \begin{IEEEbiography}[{\includegraphics[width=1in,height=1.25in,clip,keepaspectratio]{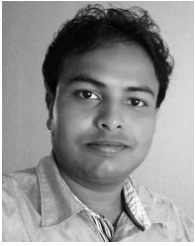}}]{Rajkumar Saini}{\space} is a university lecturer in the field of computer science. He received his Ph.D. degree from the Department of Computer Science and Engineering at the Indian Institute of Technology, Roorkee, India. Previously, he served as a Postdoctoral Researcher at the EISLAB Machine Learning, Luleå University of Technology, Sweden. His research interests encompass various areas, including computer vision, machine learning, pattern recognition, human-computer interface, brain signal analysis, and digital image processing. As a university lecturer, he is dedicated to sharing his knowledge and expertise with students and contributing to the academic community through research and teaching.
    \end{IEEEbiography}

    \begin{IEEEbiography}[{\includegraphics[width=1in,height=1.25in,clip,keepaspectratio]{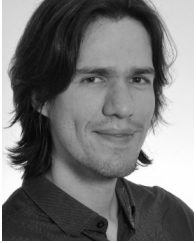}}]{Marcus Liwicki}{\space}(Member, IEEE) received the M.S.
    degree in computer science from the Free University of Berlin, Germany, in 2004, and the Ph.D.
    degree from the University of Bern, Switzerland,
    in 2007. He worked as a Senior Researcher and
    a Lecturer with the German Research Center for
    Artificial Intelligence (DFKI). He is currently a
    Professor and the Head of machine learning subject
    with the Luleå University of Technology (LTU University). His research interests include knowledge
    management, semantic desktop, electronic pen-input
    devices, online and offline handwriting recognition, and document analysis.
    He is a member of the IAPR, an Editor or a Regular Reviewer for international journals, including
    IEEE TRANSACTIONS ON PATTERN ANALYSIS AND MACHINE INTELLIGENCE, IEEE
    TRANSACTIONS ON AUDIO, SPEECH AND LANGUAGE PROCESSING, International
    Journal of Document Analysis and Recognition (an Editor), Frontiers of
    Computer Science (an Editor), Frontiers in Digital Humanities (an Editor),
    Pattern Recognition, and Pattern Recognition Letters.    
    \end{IEEEbiography}

\EOD

\end{document}